\documentclass[conference]{IEEEtran}
\IEEEoverridecommandlockouts

\usepackage{cite}
\usepackage{amsmath,amssymb,amsfonts}
\usepackage{algorithmic}
\usepackage{algorithm}
\usepackage{booktabs}
\usepackage{arydshln}
\usepackage{graphicx}
\usepackage{textcomp}
\usepackage{xcolor}
\usepackage{subcaption}
\usepackage{float} 
\usepackage[hidelinks]{hyperref}

\def\BibTeX{{\rm B\kern-.05em{\sc i\kern-.025em b}\kern-.08em
    T\kern-.1667em\lower.7ex\hbox{E}\kern-.125emX}}

\begin{document}

\title{A Graph-based Framework for Online Time Series Anomaly Detection Using Model Ensemble\\
}


\author{
\IEEEauthorblockN{
Zewei Yu\IEEEauthorrefmark{1},
Jianqiu Xu\IEEEauthorrefmark{1},
Caimin Li\IEEEauthorrefmark{2}\thanks{Corresponding author: Jianqiu Xu (jianqiu@nuaa.edu.cn). This work is supported by NSFC under Grands (U23A20296 and NO.62472217).}
}
\IEEEauthorblockA{\textit{Nanjing University of Aeronautics and Astronautics\IEEEauthorrefmark{1}, Nanjing, China}}
\IEEEauthorblockA{\textit{Nanjing Normal University\IEEEauthorrefmark{2}, Nanjing, China}}
\IEEEauthorblockA{\{yuzewei, jianqiu\}@nuaa.edu.cn\IEEEauthorrefmark{1}, 260611009@njnu.edu.cn\IEEEauthorrefmark{2}}
}

\maketitle

\begin{abstract}
With the increasing volume of streaming data in industrial systems, online anomaly detection has become a critical task. The diverse and rapidly evolving data patterns pose significant challenges for online anomaly detection. Many existing anomaly detection methods are designed for offline settings or have difficulty in handling heterogeneous streaming data effectively. This paper proposes \textbf{GDME}, an unsupervised graph-based framework for online time series anomaly detection using model ensemble. GDME maintains a dynamic model pool that is continuously updated by pruning underperforming models and introducing new ones. It utilizes a dynamic graph structure to represent relationships among models and employs community detection on the graph to select an appropriate subset for ensemble. The graph structure is also used to detect concept drift by monitoring structural changes, allowing the framework to adapt to evolving streaming data. Experiments on seven heterogeneous time series demonstrate that GDME outperforms existing online anomaly detection methods, achieving improvements of up to 24\%. In addition, its ensemble strategy provides superior detection performance compared with both individual models and average ensembles, with competitive computational efficiency.

\end{abstract}

\section{Introduction}

Time series anomaly detection aims to identify data points that significantly deviate from normal temporal patterns. It is a critical task in numerous real-world applications, including aerospace~\cite{hundman2018detecting}, server monitoring~\cite{su2019robust, ren2019time}.

In these domains, the rapid surge of high-frequency data and the growing demand for real-time monitoring have shifted the focus from offline analysis to online processing of continuous data streams. Anomaly detection on such data involves several challenges, including non-stationarity, concept drift, label scarcity, and offline model obsolescence.




Online methods that employ incremental updating have been developed~\cite{bhatia2022memstream, guha2016robust, manzoor2018xstream, pevny2016loda}, allowing models to continuously adapt to new data. However, incremental updates of a single model does not fully address the problem. As shown by the results of many comprehensive benchmark~\cite{paparrizos2022tsb, wang2024tssurvey}, there exists no single universal model that achieves optimal performance across all types of time series and anomaly patterns. Instead, certain methods perform well only on time series with specific characteristics or on particular types of anomalies~\cite{paparrizos2022tsb}.

Ensembling solutions have been proposed to overcome the limitations of single models and adapt to diverse data patterns~\cite{aggarwal2015theoretical}. However, their direct application in a streaming context is limited by two main challenges: (1) the risk of performance degradation when outputs from all detectors are aggregated indiscriminately, particularly when including underperforming detectors~\cite{rayana2016less}; and (2) the substantial computational overhead from frequently updating all base models to adapt to streaming data, which may conflict with real-time processing constraints.

A promising approach is to select a suitable subset of models from the model collection. This approach is related to \textit{Ensemble Pruning}, which improves prediction efficiency by reducing ensemble size while preserving, or sometimes enhancing, generalization performance through the exclusion of poorly performing models~\cite{caruana2006getting}. By ensembling and updating only this subset, the combined knowledge of multiple models can capture diverse patterns, mitigate the influence of weaker models, and reduce the computational cost associated with updating all models. However, the absence of ground-truth labels prevents the use of standard supervised metrics, such as F1-score, to determine an effective subset.

To address the aforementioned challenges, we propose a novel framework GDME (A \underline{G}raph-based framework for Online Time Series Anomaly \underline{D}etection using \underline{M}odel \underline{E}nsemble), consisting of two core components: model ensemble based on community detection, and concept drift detection based on graph structure changes. GDME maintains a dynamic model pool, represents model relationships through a graph, and uses community detection to select a subset for ensemble-based anomaly detection. To handle concept drift, the framework monitors graph structural changes: (1) when drift is detected, it updates the model pool by pruning underperforming models and introducing new ones (hence ``dynamic''); (2) during stable periods, GDME incrementally trains only the selected subset to ensure sustained effectiveness. This design enables adaptability to evolving data while maintaining efficiency. In summary, the contributions of this paper are:

\begin{itemize}
\item \textbf{Novel Graph-based method.} GDME leverages graph structures to enable community-based model ensemble and to detect concept drift through structural changes.
\item \textbf{Generality and extensibility.} The framework is capable of integrating a broad range of anomaly detectors and can be easily extended with new models.
\item \textbf{Demonstrated effectiveness and efficiency.} Experiments on seven heterogeneous time series show that GDME outperforms existing online anomaly detection methods by up to 24\%, and its community-based ensemble achieves superior performance with competitive efficiency.
\end{itemize}

The rest of this paper is organized as follows. Section~\ref{sec:related_work} reviews related work on deep and streaming anomaly detection. Section~\ref{sec:formulation_and_coreidea} presents the problem formulation and core idea. Section~\ref{sec:framewor_detail} details the framework, including model ensemble and concept drift handling via a dynamic graph. Section~\ref{sec:experiment} reports experiments, and Section~\ref{sec:conclusion} concludes with future directions.

\section{RELATED WORK}
\label{sec:related_work}

Recent advances in deep learning have led to the development of numerous deep anomaly detection methods based on various architectures. Earlier RNN-based methods~\cite{malhotra2015long, hundman2018detecting} summarize past information in internal memory states updated at each time step. AE-based models, such as OmniAnomaly~\cite{su2019robust}, combine RNNs with variational autoencoders to capture temporal dependencies and cross-variable correlations.

Transformer-based models have recently shown effectiveness in modeling long sequences and complex patterns. Informer~\cite{zhou2021informer} reduces the quadratic complexity of standard Transformers, while Autoformer~\cite{wu2021autoformer} and FEDformer~\cite{zhou2022fedformer} handle non-stationary data using series decomposition. PatchTST~\cite{Yuqietal-2023-PatchTST} captures local patterns by segmenting series into patches. Simple MLP-based methods, such as DLinear~\cite{zeng2023transformers}, demonstrate that carefully designed MLP architectures can also effectively model historical patterns. Convolutional methods are another widely used for anomaly detection. MICN~\cite{wang2023micn} uses multi-scale convolutions to capture features, while TimesNet~\cite{wu2023timesnet} transforms 1D series into 2D tensors to model long-term dependencies. ModernTCN~\cite{donghao2024moderntcn} further enhances TCNs for larger effective receptive fields.

In streaming anomaly detection, early methods include LODA~\cite{pevny2016loda} and xStream~\cite{manzoor2018xstream}, which build detector ensembles from random projections, with xStream further adding half-space chains for evolving features. Another common line of research is based on the isolation principle, exemplified by Random Cut Forest (RCF)~\cite{guha2016robust}, which identifies anomalies via a forest of randomly cut trees. More recently, deep models have emerged: MemStream~\cite{bhatia2022memstream} integrates a denoising autoencoder with a memory module to handle concept drift, while ARCUS~\cite{yoon2022adaptive} improves autoencoder-based detection via adaptive pooling and drift-aware updates. Beyond incremental updates, non-incremental online decision strategies have also been studied under random arrival models~\cite{tong2016online}.

\section{Problem Formulation and Core Idea}
\label{sec:formulation_and_coreidea}



\subsection{Problem Setting}

The setting of this work involves a continuous multivariate time series data stream, denoted as $\{(x^{(\tau)}, y^{(\tau)})\}_{\tau=0}^{\infty}$. For each $\tau$, a $d$-dimensional observation vector $x^{(\tau)} \in \mathbb{R}^d$ is received, accompanied by a ground-truth binary label $y^{(\tau)} \in \{0, 1\}$, where $y^{(\tau)}=1$ indicates an anomaly. The label sequence $Y$ is used solely for evaluation and comparison of method performance and is not involved in model training or ensembling.

\noindent\textbf{Definition 1 (Architecture Set).} We define an \textit{Architecture Set}, $\mathcal{A}_{\text{set}} = \{A_i\}_{i=1}^{N}$, as a repository of $N$ distinct anomaly detection architectures. Each element $A_i$ represents a unique anomaly detection algorithm (e.g., TimesNet, OmniAnomaly), rather than an instantiated model.

\noindent\textbf{Definition 2 (Model Pool).} Based on the Architecture Set, we maintain a \textit{Model Pool}, $\mathcal{M}_{\text{pool}} = \{I_j\}_{j=1}^{M}$, containing concrete models actively managed and utilized by the framework, where $M$ is the number of models in the pool. Each model $I_j \in \mathcal{M}_{\text{pool}}$ is a trained detector represented by a triplet $(A, H, \theta)$, where $A \in \mathcal{A}_{\text{set}}$ denotes an anomaly detection algorithm, $H$ specifies a hyperparameter configuration, and $\theta$ represents the parameters learned during training. For example, one model could be $(\text{OmniAnomaly}, \{\text{hidden\_channels}=38, \text{num\_layers}=2, \dots\}, \theta)$. 

For processing, the stream is handled in a batch-wise manner. Let the stream be represented as $DS = \{B^{(t)}\}_{t=0}^{\infty}$. Models are instantiated from $\mathcal{A}_{\text{set}}$ and trained on the initial batch $B_0$ to form the initial model pool $\mathcal{M}_{\text{pool}}^{(0)}$.

For each newly arrived batch $B^{(t)}$ with $t>0$, models in $\mathcal{M}_{\text{pool}}^{\boldsymbol{(t-1)}}$ first produce anomaly scores for evaluation, and are then updated on the same batch following the prequential evaluation scheme~\cite{gama2013evaluating}, resulting in the updated model pool $\mathcal{M}_{\text{pool}}^{(t)}$. Let $\boldsymbol{S}^{(t)} = \{\boldsymbol{s}_1^{(t)}, \dots, \boldsymbol{s}_M^{(t)}\}$ denote the anomaly score set, where $\boldsymbol{s}_j^{(t)}$ contains the scores from model $I_j^{(t-1)}$ on $B^{(t)}$. This set is used to construct the model graph $G^{(t)}$ for ensemble and concept drift detection. The details are provided in Section~\ref{sec:framewor_detail}.

\subsection{Core Idea}
\label{sec:core_idea}
To address the challenges in online time series anomaly detection, an promising strategy is to select a representative subset of models from a model pool for ensembling. Intuitively, if a group of models exhibits similar anomaly detection behavior on the same batch, retaining only one as a representative is sufficient, which naturally corresponds to clustering. However, this idea faces two fundamental challenges:

\begin{itemize}
    \item \textbf{Representation problem:} Clustering requires each sample to be a fixed-dimensional vector, but models differ in architecture and parameters, making universal vectorization difficult. 

    \item \textbf{Representative selection problem:} Even if clustering is feasible, choosing a representative from each cluster is difficult in an unsupervised setting, where ground-truth labels are unavailable.
\end{itemize}

To overcome these challenges, we propose a paradigm shift: \textbf{focusing on external behavioral relationships between models rather than their internal parameters}. Specifically, we characterize model behavior by the correlations between their anomaly score sequences, and represent the model pool as a dynamic graph, where nodes denote models and edge weights capture behavioral similarity. Communities in this graph correspond to coherent groups of models, and representative models can be selected using unsupervised metrics such as node centrality.

\section{Proposed Framework}
\label{sec:framewor_detail}

\subsection{Overview}
\begin{figure}[H]
    \centering
    \includegraphics[width=0.9\linewidth]{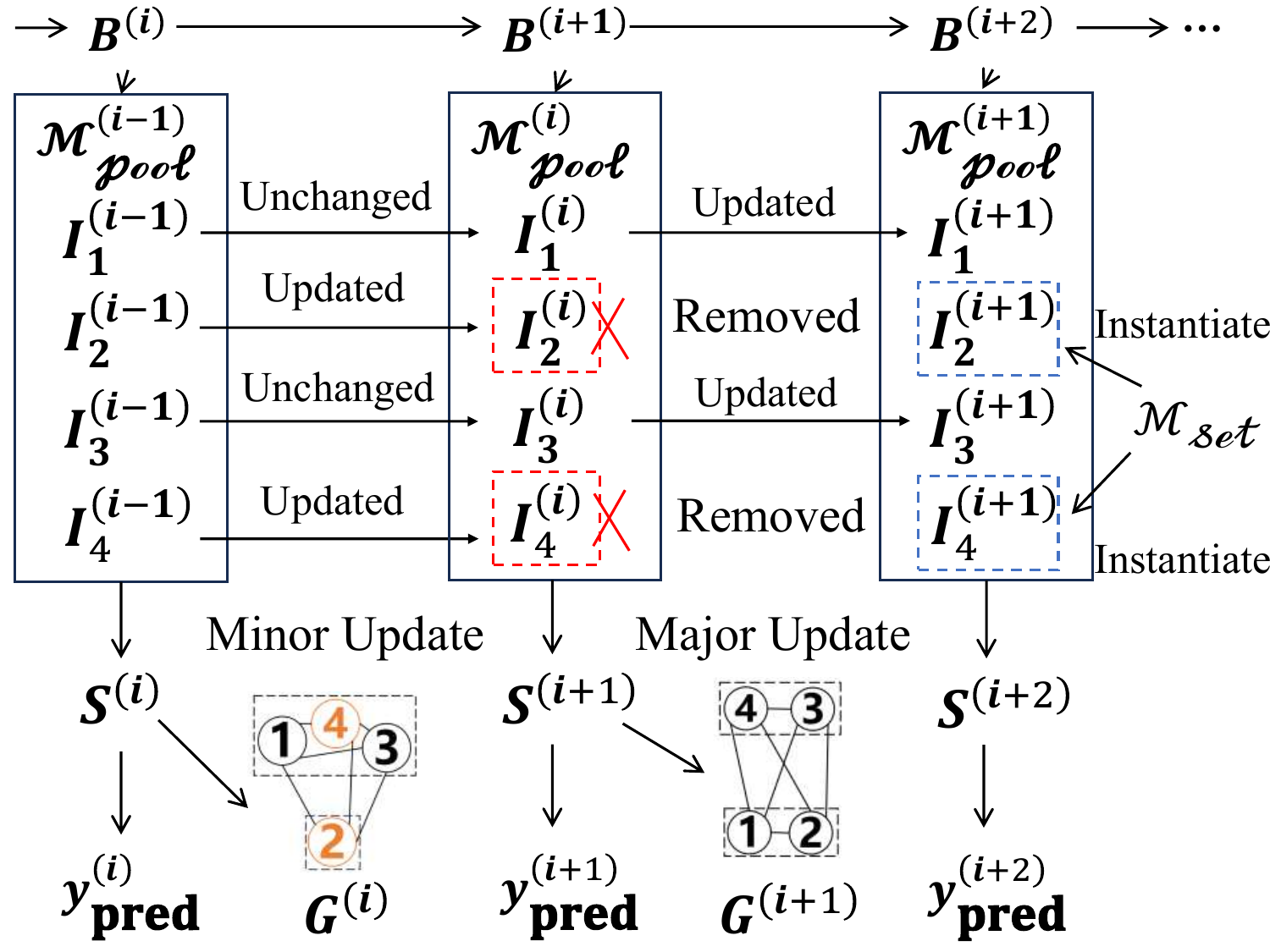}
    \caption{Illustration of the GDME Framework.}
    \label{fig:stream_processing}
\end{figure}

GDME is a model ensemble framework for online time series anomaly detection (Algorithm~\ref{alg:GDME_overview}, Fig.~\ref{fig:stream_processing}). The notations and functions used in Algorithm~\ref{alg:GDME_overview} are summarized in Table~\ref{tab:GDME_notion} and Table~\ref{tab:GDME_functions}, respectively. GDME maintains a dynamic model pool, $\mathcal{M}_{\text{pool}}$, and represents relationships among models using a graph structure. During initialization, models are instantiated from the Architecture Set $\mathcal{A}_{\text{set}}$ and trained on the initial batch $B_0$ to form $\mathcal{M}_{\text{pool}}^{(0)}$ (Line 1). 

\begin{table}[htbp]
\centering
\caption{Notation used in Algorithm~\ref{alg:GDME_overview}.}
\label{tab:GDME_notion}
\begin{tabular}{p{0.2\columnwidth} p{0.7\columnwidth}}
\toprule
\textbf{Symbol} & \textbf{Description} \\
\midrule
$DS = \{B^{(t)}\}$ & Data stream, $B^{(t)}$ is the $t$-th batch. \\
$\mathcal{A}_{\text{set}}$ & Architecture set of heterogeneous anomaly detectors. \\
$\theta_{\text{drift}}$ & Threshold for drift detection. \\
$\boldsymbol{Y}_{\text{pred}}$ & Predicted result for the entire data stream. \\
$\mathcal{M}_{\text{pool}}^{(t-1)}$ & Model pool at time $t-1$, to be used for $B^{(t)}$. \\
$\boldsymbol{S}^{(t)}$ & Anomaly score set of $\mathcal{M}_{\text{pool}}^{(t-1)}$ on $B^{(t)}$. \\
$\boldsymbol{C}^{(t)}$ & Correlation matrix of $\boldsymbol{S}^{(t)}$. \\
$V^{(t)}$ & Node set at time $t$, each representing a model. \\
$E^{(t)}$ & Edge set at time $t$, each weight encoding correlation between two models. \\
$G^{(t)}$ & Model graph at time $t$, built from $V^{(t)}$ and $E^{(t)}$. \\
$\mathcal{P}^{(t)}$ & Partition of $G^{(t)}$ via community detection. \\
$\mathcal{R}^{(t)}$ & Representatives selected from $\mathcal{P}^{(t)}$. \\
$\boldsymbol{s}_{\text{final}}^{(t)}$ & Final ensemble anomaly score. \\
$D^{(t)}$ & Drift score between $G^{(t)}$ and $G^{(t-1)}$. \\
$\boldsymbol{y}_{\text{pred}}^{(t)}$ & Predicted anomaly labels for $B^{(t)}$. \\
\bottomrule
\end{tabular}
\end{table}

\begin{table}[htbp]
\centering
\caption{Functions used in Algorithm~\ref{alg:GDME_overview}.}
\label{tab:GDME_functions}
\begin{tabular}{p{0.35\columnwidth} p{0.55\columnwidth}}
\toprule
\textbf{Function} & \textbf{Description} \\
\midrule
\texttt{train($\cdot, B$)} & Train architectures or models specified by the first argument on batch $B$. \\
\texttt{score($\mathcal{M}_{\text{pool}}, B$)} & Compute anomaly scores for $\mathcal{M}_{\text{pool}}$ on batch $B$. \\
\texttt{corr($\boldsymbol{S}$)} & Compute pairwise rank correlation between score vectors in score set $\boldsymbol{S}$. \\
\texttt{community\_detect($G$)} & Perform community detection on graph $G$, returning partition of model clusters. \\
\texttt{select\_rep($\mathcal{C}$)} & Select a representative model from community $\mathcal{C}$. \\
\texttt{ensemble($\mathcal{R}, \boldsymbol{S}$)} & Aggregate anomaly scores from representative models $\mathcal{R}$ into a final score. \\
\texttt{drift\_score($G_1, G_2$)} & Compute drift score between two graphs $G_1$ and $G_2$. \\
\texttt{prune($\mathcal{M}_{\text{pool}}$)} & Remove underperforming models from $\mathcal{M}_{\text{pool}}$. \\
\texttt{threshold($\boldsymbol{s}$)} & Convert score vector $\boldsymbol{s}$ into binary predictions. \\
\bottomrule
\end{tabular}
\end{table}

\begin{algorithm}[htbp]
\caption{GDME}
\label{alg:GDME_overview}
\begin{algorithmic}[1]
\REQUIRE $DS = \{B^{(t)}\}_{t=0}^{\infty}$, $\mathcal{A}_{\text{set}}$, $\theta_{\text{drift}}$
\ENSURE $\boldsymbol{Y}_{\text{pred}}$

\STATE $\mathcal{M}_{\text{pool}}^{(0)} \gets \text{train}(\mathcal{A}_{\text{set}}, B_0)$
\STATE $\boldsymbol{Y}_{\text{pred}} \gets []$
\FOR{$t = 1,2,\dots$}
    \STATE $\boldsymbol{S}^{(t)} \gets \text{score}(\mathcal{M}_{\text{pool}}^{(t-1)}, B^{(t)})$
    \item[] \textcolor{blue}{// Graph Construction}
    \STATE $\boldsymbol{C}^{(t)} \gets \text{corr}(\boldsymbol{S}^{(t)})$
    \STATE $V^{(t)} \gets \mathcal{M}_{\text{pool}}^{(t-1)}$
    \STATE $E^{(t)} \gets \{(i,j,w_{ij}=\boldsymbol{C}^{(t)}_{ij}) \mid i,j\in V^{(t)}\}$
    \STATE $G^{(t)} \gets (V^{(t)}, E^{(t)})$
    \item[] \textcolor{blue}{// Graph Community-based Model Ensemble}
    \STATE $\mathcal{P}^{(t)} \gets \text{community\_detect}(G^{(t)})$
    \STATE $\mathcal{R}^{(t)} \gets \{ \text{select\_rep}(\mathcal{C}) \mid \mathcal{C} \in \mathcal{P}^{(t)} \}$
    \STATE $\boldsymbol{s}_{\text{final}}^{(t)} \gets \text{ensemble}(\mathcal{R}^{(t)}, \boldsymbol{S}^{(t)})$
    \item[] \textcolor{blue}{// Graph-based Concept Drift Detection}
    \STATE $D^{(t)} \gets \text{drift\_score}(G^{(t)}, G^{(t-1)})$

    \IF{$D^{(t)} > \theta_{\text{drift}}$}
        \item[] \textcolor{blue}{// Drift Detected: Major Update}
        \STATE prune$(\mathcal{M}_{\text{pool}}^{(t-1)})$
        \STATE train$(\mathcal{M}_{\text{pool}}^{(t-1)}, B^{(t)})$
        \STATE $\mathcal{M}_{\text{pool}}^{(t)} \gets \mathcal{M}_{\text{pool}}^{(t-1)} \cup \text{train}(\mathcal{A}_{\text{set}}, B^{(t)})$
    \ELSE
        \item[] \textcolor{blue}{// No Drift: Minor Update}
        \STATE train$(\mathcal{R}^{(t)}, B^{(t)})$
        \STATE $\mathcal{M}_{\text{pool}}^{(t)} \gets \mathcal{M}_{\text{pool}}^{(t-1)}$
    \ENDIF

    \STATE $\boldsymbol{y}_{\text{pred}}^{(t)} \gets \text{threshold}(\boldsymbol{s}_{\text{final}}^{(t)})$
    \STATE Append $\boldsymbol{y}_{\text{pred}}^{(t)}$ to $\boldsymbol{Y}_{\text{pred}}$

    \IF{any($\boldsymbol{y}_{\text{pred}}^{(t)}=1$)}
        \STATE trigger alarm
    \ENDIF
\ENDFOR
\RETURN $\boldsymbol{Y}_{\text{pred}}$
\end{algorithmic}
\end{algorithm}

For each batch $B^{(t)}$ ($t \ge 1$), anomaly scores $\boldsymbol{S}^{(t)}$ are first computed for all models in $\mathcal{M}_{\text{pool}}^{(t-1)}$ (Line 4). The pairwise rank correlation matrix $\boldsymbol{C}^{(t)}$ is then calculated to define the weighted edges, which together with the nodes construct the model graph $G^{(t)}$ (Line 5–8). Graph community detection partitions the models into groups $\mathcal{P}^{(t)}$, and one representative from each community is selected to form the ensemble subset $\mathcal{R}^{(t)}$, whose scores are averaged to obtain the final anomaly score $\boldsymbol{s}_{\text{final}}^{(t)}$ (Line 9–11). To address concept drift, the framework computes a drift score $D^{(t)}$ from topological changes in $G^{(t)}$ relative to $G^{(t-1)}$ (Line 12). If $D^{(t)}$ exceeds the drift threshold, a major update is performed: underperforming models are pruned, the remaining models are incrementally updated on $B^{(t)}$, new models are instantiated from $\mathcal{A}_{\text{set}}$ and trained on $B^{(t)}$ to form $\mathcal{M}_{\text{pool}}^{(t)}$ (Line 14–16). Otherwise, only the selected subset $\mathcal{R}^{(t)}$ is incrementally trained, and the model pool remains unchanged (Line 18-19). The final anomaly scores $\boldsymbol{s}_{\text{final}}^{(t)}$ are converted into binary predictions $\boldsymbol{y}_{\text{pred}}^{(t)}$, and these predictions are appended to the accumulated list $\boldsymbol{Y}_{\text{pred}}$ (Line 21–22). An anomaly alarm is triggered if any positive prediction is detected in the current batch (Line 24). Finally, the algorithm returns $\boldsymbol{Y}_{\text{pred}}$ (Line 27).

\subsection{Graph Construction Method}
Given the anomaly score set $\boldsymbol{S}^{(t)} = \{\boldsymbol{s}^{(t)}_{1}, \boldsymbol{s}^{(t)}_{2}, \dots, \boldsymbol{s}^{(t)}_{M}\}$, we first construct the adjacency matrix $\mathbf{C}^{(t)} = [w_{ij}^{(t)}]$, where each entry encodes the pairwise behavioral consistency between models.  
Formally, for two models $I_i^{(t-1)}$ and $I_j^{(t-1)}$, the edge weight is defined as
\begin{equation}
w_{ij}^{(t)} = w_{ji}^{(t)} = 
\rho \big(\boldsymbol{s}_i^{(t)}, \boldsymbol{s}_j^{(t)} \big), \quad i \neq j.
\label{eq:edge_weight}
\end{equation}
where $\rho(\cdot,\cdot)$ denotes a correlation metric.
Specifically, Spearman’s rank correlation coefficient is adopted, as it depends solely on the relative ranking of anomaly scores and is independent of their absolute values. Self-correlation is excluded by setting diagonal entries of $\mathbf{C}^{(t)}$ to zero, preventing self-loops in the resulting graph.  

The undirected weighted graph $G^{(t)} = (V^{(t)}, E^{(t)}, w^{(t)})$ is then constructed from $\mathbf{C}^{(t)}$, where $V^{(t)}$ is the set of models, $E^{(t)} = \{(i,j) \mid w_{ij}^{(t)} \neq 0\}$ is the edge set, and $w^{(t)}$ is the edge-weight function given by the entries of $\mathbf{C}^{(t)}$. By updating $\mathbf{C}^{(t-1)}$ to $\mathbf{C}^{(t)}$ using batch $B^{(t)}$, the framework maintains a dynamic graph sequence $\{G^{(t)}\}_{t=1}^{\infty}$ that captures the evolving relational structure of the model pool.

\subsection{Graph Community-based Model Ensemble}
\label{sec:model_selection_for_ensemble}

After constructing the graph $G^{(t)} = (V^{(t)}, E^{(t)}, w^{(t)})$, nodes are partitioned into disjoint communities $\mathcal{P}^{(t)} = \{\mathcal{C}_1^{(t)}, \dots, \mathcal{C}_k^{(t)}\}$, with each $\mathcal{C}_i^{(t)} \subseteq V^{(t)}$ representing a subset of models. Weighted community detection maximizes intra-community edge weights and minimizes inter-community edge weights, grouping models with strongly correlated anomaly scores into the same community. The Louvain method~\cite{blondel2008fast} is used and its resolution parameter is discussed in Section~\ref{sec:louvain_influence}.

From each community $\mathcal{C}_i^{(t)}$, a single representative node is selected, forming the representative subset $\mathcal{R}^{(t)} \subseteq \mathcal{M}_\text{pool}^{(t-1)}$. Since models within the same community have highly correlated anomaly scores, retaining only one avoids redundancy. Existing selection strategies include selecting the model with lowest reconstruction error or highest centrality~\cite{goswami2022unsupervised}. Here, a two-level approach combining graph-based centrality and an unsupervised pseudo-performance score is used to select representatives.

Specifically, let $\mathcal{G}_i^{(t)} := G^{(t)}[\mathcal{C}_i^{(t)}]$ denote the subgraph induced by the nodes in community $\mathcal{C}_i^{(t)}$. For each node $v_j^{(t)} \in \mathcal{C}_i^{(t)}$, its PageRank centrality is computed as
\begin{equation}
c_j^{(t)} = \text{PageRank}(\mathcal{G}_i^{(t)})[v_j^{(t)}],
\end{equation}
where PageRank$(\cdot)$ returns a mapping from each node to its centrality score. Centrality alone may not reflect detection performance, especially in communities with many poorly performing models. To address this, a pseudo-ground truth approach~\cite{rayana2016less} is adopted. Each score vector in $\boldsymbol{S}^{(t)} = \{\boldsymbol{s}^{(t)}_{1}, \dots, \boldsymbol{s}^{(t)}_{M}\}$ is binarized via a Gaussian Mixture model to separate inliers and outliers. In our implementation, we use a fixed configuration with two components. The resulting labels are aggregated by majority voting to form a pseudo-ground truth $\tilde{y}^{(t)}$. Each node $v_j^{(t)} \in \mathcal{C}_i^{(t)}$ is then assigned a pseudo-performance score $q_j^{(t)}$ by computing the AUC between its score vector $\boldsymbol{s}_j^{(t)}$ and the pseudo-ground truth $\tilde{y}^{(t)}$. The final score for selecting a representative model combines normalized centrality and pseudo-performance:

\begin{equation}
h_j^{(t)} = \alpha c_j^{(t)} + (1-\alpha) q_j^{(t)},
\end{equation}
where $\alpha \in [0,1]$ balances the contributions of the two terms.

The representative node for community $\mathcal{C}_i^{(t)}$ is selected as the one with the highest combined score $h_j^{(t)}$, which reflects both centrality and pseudo-performance. Representatives from all communities form $\mathcal{R}^{(t)}$, used in an average ensemble to produce the final anomaly score $\boldsymbol{s}_{\text{final}}^{(t)}$ for batch $B^{(t)}$.

\subsection{Graph-based Concept Drift Detection}
\label{sec:graph_based_concept_drift_detection}

Concept drift is hypothesized to induce changes in both the community structure and the distribution of node centrality values in the graph. For batch $B^{(t)}$, the framework compares the current partition $\mathcal{P}^{(t)}$ and the ranking of node centrality values with those from the previous batch, $\mathcal{P}^{(t-1)}$, and the previous centrality ranking. These comparisons yield two complementary drift measures: \textit{centrality drift}, capturing changes in node centrality, and \textit{community drift}, capturing changes in community structure.

The ranking of node centrality values is computed by first calculating the centrality of the nodes on the entire graph $G^{(t)}$, which are then converted into a ranking vector $\boldsymbol{CR}^{(t)}$, which reflects the relative centrality of the nodes within the graph. Centrality drift is quantified by comparing the rankings between consecutive batches using Kendall's $\tau$, denoted as $d_\text{cent}^{(t)}$:
\begin{equation}
d_\text{cent}^{(t)} = \frac{1 - \tau(\boldsymbol{CR}^{(t-1)}, \boldsymbol{CR}^{(t)})}{2},
\end{equation}
where $\tau(\cdot,\cdot) \in [-1,1]$ is Kendall's rank correlation coefficient, which measures the ordinal association between two rankings, with smaller $\tau$ indicating larger changes in centrality order.

Community drift is quantified by comparing the partitions of nodes at $\mathcal{P}^{(t-1)}$ and $\mathcal{P}^{(t)}$, denoted as $d_\text{comm}^{(t)}$:
\begin{equation}
d_\text{comm}^{(t)} = 1 - \text{NMI}(\mathcal{P}^{(t-1)}, \mathcal{P}^{(t)}),
\end{equation}
where $\text{NMI}(\cdot,\cdot)$ denotes the normalized mutual information between two partitions, taking values in $[0,1]$, with lower NMI indicating greater drift. NMI is a metric that quantifies the similarity between two partitions.

The overall drift score is obtained as a weighted combination of the two signals:
\begin{equation}
D^{(t)} = \beta \, d_\text{comm}^{(t)} + (1-\beta) \, d_\text{cent}^{(t)},
\end{equation}
where $\beta \in [0,1]$ balances the contributions of community and centrality drift. Concept drift is detected if $D^{(t)} > \theta_{\text{drift}}$.

\subsection{Model Pool Update Strategy}

The model pool is continuously updated as new data batches arrive, performing a major update if concept drift is detected, and a minor update otherwise. Each model is assigned a \textit{contribution score} reflecting its role as a representative on past data. For each normal (no-drift) batch $B^{(t)}$ occurring after the most recent detected concept drift, applying the representative selection method in Section~\ref{sec:model_selection_for_ensemble} yields a set of representative models $\mathcal{R}^{(t)}$. A counter continuously tracks how many times each model has been selected as a representative since the latest drift. For efficiency, only the selected representatives are incrementally trained on the new batch.

When concept drift is detected at batch $B^{(t+l)}$ ($l>0$), it indicates that all models in $\mathcal{M}_\text{pool}^{(t+l-1)}$ are no longer sufficient and new models need to be added. To ensure efficiency, the pool has a maximum capacity $N_\text{max}$. New models are instantiated from all architectures in $\mathcal{A}_\text{set}$, and if adding new models causes the pool size to exceed $N_\text{max}$, some existing models must be pruned first. Let $n_j^{(t)}$ be the count of times model $I_j$ was selected as a representative in stable batches since the last detected drift. Its short-term contribution score is defined as
\begin{equation}
cs_j^{(t)} = \frac{n_j^{(t)}}{\sum_{k} n_k^{(t)}}.
\end{equation}
where the denominator sums over all models and the counters $n_j^{(t)}$ are reset afterward (hence ``short-term''). The long-term contribution scores of models added to model pool at the last drift are initialized as $CS_j^{(t+l)} = cs_j$, while for all other models, the long-term scores are updated via an exponential moving average (hence ``long-term'').
\begin{equation}
CS_j^{(t+l)} = \gamma \, cs_j + (1-\gamma) \, CS_j^{(t_\text{prev})},
\end{equation}
where $CS_j^{(t_\text{prev})}$ denotes the long-term contribution score before the current drift, and $\gamma \in [0,1]$ controls the weight of the most recent contribution.

If adding new models were to exceed the pool's maximum capacity $N_\text{max}$, pruning would be applied. Let
\begin{equation}
n_\text{exceed} = \max\big(0, |\mathcal{M}_\text{pool}^{(t+l-1)}| + |\mathcal{A}_{\text{set}}| - N_\text{max}\big)
\end{equation}
be the number of models to remove. The $n_\text{exceed}$ models with the lowest long-term contribution scores are pruned.

Finally, all existing models in $\mathcal{M}_\text{pool}^{(t+l-1)}$ are incrementally trained on $B^{(t+l)}$, while new models instantiated from $\mathcal{A}_{\text{set}}$ are trained on the same batch, assigned undefined long-term scores, and added to the pool, forming $\mathcal{M}_\text{pool}^{(t+l)}$.

\section{EXPERIMENTAL EVALUATION}
\label{sec:experiment}

\begin{table*}[ht]
\centering
\caption{Comparison of GDME and online anomaly detection baselines in terms of AUC and ADT (ms).}
\label{tab:baseline_cmp}
\begin{tabular*}{\textwidth}{@{\extracolsep{\fill}} l cc cc cc cc cc cc cc}
\toprule
\multicolumn{1}{c}{} 
  & \multicolumn{2}{c}{CalIt2} 
  & \multicolumn{2}{c}{Dodgers} 
  & \multicolumn{2}{c}{MSL} 
  & \multicolumn{2}{c}{IOPS\_1c6} 
  & \multicolumn{2}{c}{IOPS\_05f} 
  & \multicolumn{2}{c}{SMD\_1\_1} 
  & \multicolumn{2}{c}{SMD\_3\_7} \\
\multicolumn{1}{c}{} 
  & \multicolumn{1}{c}{AUC} & \multicolumn{1}{c}{ADT} 
  & \multicolumn{1}{c}{AUC} & \multicolumn{1}{c}{ADT} 
  & \multicolumn{1}{c}{AUC} & \multicolumn{1}{c}{ADT} 
  & \multicolumn{1}{c}{AUC} & \multicolumn{1}{c}{ADT} 
  & \multicolumn{1}{c}{AUC} & \multicolumn{1}{c}{ADT} 
  & \multicolumn{1}{c}{AUC} & \multicolumn{1}{c}{ADT} 
  & \multicolumn{1}{c}{AUC} & \multicolumn{1}{c}{ADT} \\
\midrule
LODA~\cite{pevny2016loda}            & 0.534       & 2.375        & 0.624        & 2.281        & 0.616         & 2.526         & 0.553         & 2.261         & 0.523         & 2.329         & 0.526        & 2.464          & 0.504        & 2.550          \\
RCF~\cite{guha2016robust}            & 0.684       & 47.117       & 0.668        & 31.426       & 0.575         & 1558.920        & 0.896         & 37.189        & 0.932         & 36.915        & 0.750        & 336.558        & 0.727        & 394.463        \\
xStream~\cite{manzoor2018xstream}    & 0.608       & 20.221       & 0.671        & 17.921       & 0.548         & 203.785        & 0.579         & 17.657        & 0.548         & 17.660        & 0.549        & 145.406        & 0.515        & 144.384        \\
MemStream~\cite{bhatia2022memstream} & 0.695       & 0.189        & 0.579        & 0.170        & 0.586         & 0.503         & 0.919         & 0.174         & 0.871         & 0.174         & 0.588        & 5.174          & \textbf{0.924}        & 5.878          \\
\textbf{GDME-CPU}                        & 0.863       & 6.365        & 0.683        & 5.863        & 0.610         & 44.373         & 0.930         & 6.958         & 0.949         & 8.117         & 0.855        & 27.472         & 0.824        & 29.222         \\ 
\textbf{GDME}                        & \textbf{0.866}       & 2.632        & \textbf{0.705}        & 3.106        & \textbf{0.649}         & 3.628         & \textbf{0.934}         & 2.767         & \textbf{0.963}         & 2.859         & \textbf{0.863}        & 5.473          & 0.862        & 5.862          \\
\bottomrule
\end{tabular*}
\end{table*}

\begin{table*}[h]
\centering
\caption{Comparison of individual methods, the average ensemble, and GDME in terms of AUC and ADT (ms).}
\label{tab:ensemble_comparison}
\begin{tabular*}{\textwidth}{@{\extracolsep{\fill}} l|cc|cc|cc|cc|cc|cc|cc}
\toprule
\multicolumn{1}{c}{} 
  & \multicolumn{2}{c}{CalIt2} 
  & \multicolumn{2}{c}{Dodgers} 
  & \multicolumn{2}{c}{MSL} 
  & \multicolumn{2}{c}{IOPS\_1c6} 
  & \multicolumn{2}{c}{IOPS\_05f} 
  & \multicolumn{2}{c}{SMD\_1\_1} 
  & \multicolumn{2}{c}{SMD\_3\_7} \\
\multicolumn{1}{c}{} 
  & \multicolumn{1}{c}{AUC} & \multicolumn{1}{c}{ADT} 
  & \multicolumn{1}{c}{AUC} & \multicolumn{1}{c}{ADT} 
  & \multicolumn{1}{c}{AUC} & \multicolumn{1}{c}{ADT} 
  & \multicolumn{1}{c}{AUC} & \multicolumn{1}{c}{ADT} 
  & \multicolumn{1}{c}{AUC} & \multicolumn{1}{c}{ADT} 
  & \multicolumn{1}{c}{AUC} & \multicolumn{1}{c}{ADT} 
  & \multicolumn{1}{c}{AUC} & \multicolumn{1}{c}{ADT} \\
\midrule
OmniAnomaly~\cite{su2019robust}         & 0.694            & 0.058   & 0.625             & 0.055   & 0.636              & 0.122    & 0.735              & 0.052    & 0.817              & 0.049    & 0.751             & 0.100     & 0.675             & 0.100     \\
Informer~\cite{zhou2021informer}        & 0.763            & 0.096   & 0.538             & 0.105   & 0.627              & 0.167    & 0.815              & 0.102    & 0.840              & 0.099    & 0.771             & 0.140     & 0.832             & 0.139     \\
Autoformer~\cite{wu2021autoformer}      & 0.813            & 0.155   & 0.627             & 0.153   & 0.624              & 0.238    & 0.721              & 0.142    & 0.828              & 0.134    & 0.788             & 0.196     & 0.693             & 0.198     \\
FEDformer~\cite{zhou2022fedformer}      & 0.772            & 1.044   & 0.654             & 0.978   & \textbf{0.651}              & 1.027    & 0.729              & 0.949    & 0.843              & 1.070    & 0.787             & 1.000     & 0.683             & 0.955     \\
Crossformer~\cite{zhang2023crossformer} & 0.832            & 0.112   & 0.554             & 0.152   & 0.618              & 0.487    & 0.874              & 0.141    & 0.902              & 0.134    & 0.608             & 0.338     & 0.725             & 0.337     \\
PatchTST~\cite{Yuqietal-2023-PatchTST}  & 0.812            & 0.072   & 0.608             & 0.075   & 0.634              & 0.495    & 0.901              & 0.071    & 0.864              & 0.073    & 0.848             & 0.340     & 0.678             & 0.338     \\
iTransformer~\cite{liu2024itransformer} & 0.830            & 0.068   & 0.606             & 0.063   & 0.607              & 0.127    & 0.902              & 0.062    & 0.881              & 0.062    & 0.709             & 0.108     & 0.718             & 0.103     \\
LightTS~\cite{Zhang2022LessIM}          & 0.791            & 0.049   & 0.577             & 0.047   & 0.606              & 0.101    & 0.865              & 0.043    & 0.876              & 0.043    & 0.793             & 0.081     & 0.737             & 0.084     \\
DLinear~\cite{zeng2023transformers}     & 0.847            & 0.034   & 0.534             & 0.014   & 0.622              & 0.294    & 0.883              & 0.014    & 0.874              & 0.016    & 0.731             & 0.209     & 0.730             & 0.229     \\
TSMixer~\cite{chen2023tsmixer}          & 0.812            & 0.037   & 0.554             & 0.030   & 0.595              & 0.093    & 0.867              & 0.029    & 0.879              & 0.030    & 0.737             & 0.071     & 0.791             & 0.076     \\
MTS-Mixers~\cite{Li2023MTSMixersMT}     & 0.784            & 0.042   & 0.644             & 0.028   & 0.633              & 0.227    & 0.864              & 0.027    & 0.888              & 0.028    & 0.786             & 0.174     & 0.752             & 0.181     \\
SCINet~\cite{liu2022scinet}             & 0.837            & 0.095   & 0.606             & 0.113   & 0.600              & 0.174    & 0.896              & 0.112    & 0.869              & 0.113    & 0.797             & 0.146     & 0.745             & 0.147     \\
MICN~\cite{wang2023micn}                & 0.731            & 0.073   & 0.540             & 0.078   & 0.577              & 0.140    & 0.905              & 0.074    & 0.884              & 0.075    & 0.696             & 0.109     & 0.651             & 0.111     \\
TimesNet~\cite{wu2023timesnet}          & 0.832            & 1.115   & 0.620             & 1.121   & 0.581              & 3.460    & 0.897              & 1.118    & 0.882              & 1.119    & 0.846             & 3.374     & 0.815             & 3.337     \\
ModernTCN~\cite{donghao2024moderntcn}   & 0.768            & 0.047   & 0.625             & 0.032   & 0.624              & 0.194    & 0.862              & 0.033    & 0.890              & 0.033    & 0.783             & 0.125     & 0.716             & 0.128     \\
Average Ensemble                        & 0.849            & 8.758   & 0.680             & 8.763   & 0.629              & 11.401    & 0.907              & 8.716    & 0.944              & 9.140    & 0.845             & 16.520    & 0.796             & 16.280    \\
\textbf{GDME}                             & \textbf{0.866}   & 2.632   & \textbf{0.705}    & 3.106   & 0.649     & 3.628    & \textbf{0.934}     & 2.767    & \textbf{0.963}     & 2.859    & \textbf{0.863}    & 5.473     & \textbf{0.862}    & 5.862     \\
\bottomrule
\end{tabular*}
\end{table*}

\subsection{Experimental Setup}

\subsubsection{Datasets}

We conducted experiments on seven heterogeneous time series from four real-world datasets, including CalIt2~\cite{ihler2006adaptive}, Dodgers~\cite{ihler2006adaptive}, IOPS~\cite{ren2019time}, SMD~\cite{su2019robust} and MSL~\cite{hundman2018detecting}, covering both univariate and multivariate cases. The dataset statistics are summarized in Table~\ref{tab:datasets}, where $L$ denotes the sequence length, $D$ the dimensionality, and $M$ the median anomaly length.

\begin{table}[H]
\centering
\caption{Statistics of the datasets.}
\label{tab:datasets}
\begin{tabular*}{0.9\columnwidth}{@{\extracolsep{\fill}}lccc}
\toprule
Time Series & $L$ & $D$ & $M$ \\
\midrule
CalIt2     & 5k   & 2  & 7   \\ 
Dodgers    & 50k  & 1  & 33  \\ 
MSL  & 73k  & 55 & 216  \\ 
IOPS\_1c6 & 149k & 1  & 11  \\ 
IOPS\_05f & 146k & 1  & 18  \\ 
SMD\_1\_1  & 28k  & 38 & 433 \\ 
SMD\_3\_7  & 28k  & 38 & 35  \\ 

\bottomrule
\end{tabular*}
\end{table}

\subsubsection{Algorithms}
In our experiments, the Architecture Set $\mathcal{A}_{\text{set}}$ comprises a diverse collection of deep anomaly detection algorithms, grouped into four categories: (1) AE-based: OmniAnomaly~\cite{su2019robust}; (2) Transformer-based: Informer~\cite{zhou2021informer}, Autoformer~\cite{wu2021autoformer}, FEDformer~\cite{zhou2022fedformer}, Crossformer~\cite{zhang2023crossformer}, PatchTST~\cite{Yuqietal-2023-PatchTST}, iTransformer~\cite{liu2024itransformer}; (3) MLP-based: LightTS~\cite{Zhang2022LessIM}, DLinear~\cite{zeng2023transformers}, TSMixer~\cite{chen2023tsmixer}, MTS-Mixers~\cite{Li2023MTSMixersMT}; (4) CNN-based: SCINet~\cite{liu2022scinet}, MICN~\cite{wang2023micn}, TimesNet~\cite{wu2023timesnet}, ModernTCN~\cite{donghao2024moderntcn}. In addition, we included several online anomaly detection baselines: RCF~\cite{guha2016robust}, MemStream~\cite{bhatia2022memstream}, LODA~\cite{pevny2016loda}, and xStream~\cite{manzoor2018xstream}.

\subsubsection{Metrics}

The evaluation considers two aspects: model performance and detection time. Performance is measured by the threshold-independent AUC. Detection time is quantified by the \textit{Average Detection Time for Each Time Step} (\textit{ADT}), which reflects the method's real-time processing capability. Specifically, \textit{ADT} is defined as
\begin{equation}
\textit{ADT} = \frac{\text{sum of detection times at all time steps}}{\text{total number of time steps}}.
\end{equation}



\begin{figure*}[htbp]
    \centering

    \begin{subfigure}[b]{0.6\textwidth}
        \centering
        \includegraphics[width=\textwidth]{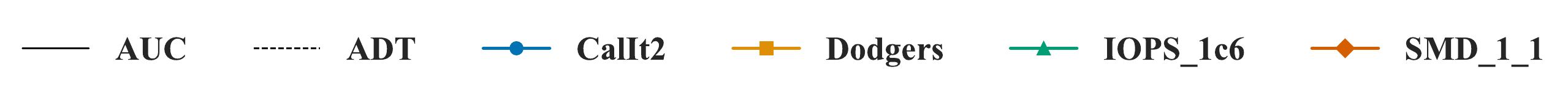}
    \end{subfigure}
    
    \begin{subfigure}[b]{0.23\textwidth}
        \centering
        \includegraphics[width=\textwidth]{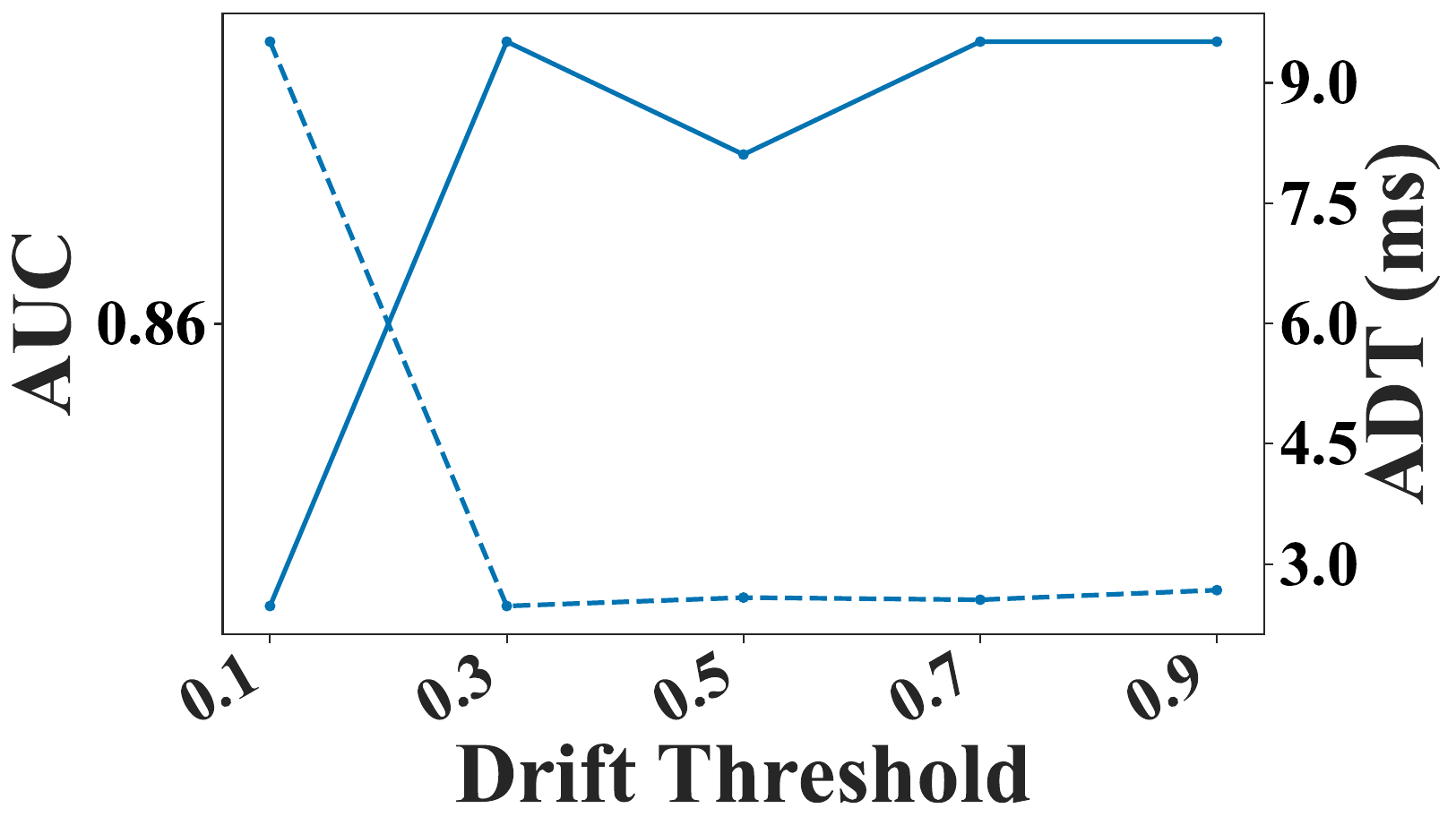}
        \caption{CalIt2.}
        \label{fig:calit2_drift}
    \end{subfigure}
    \begin{subfigure}[b]{0.23\textwidth}
        \centering
        \includegraphics[width=\textwidth]{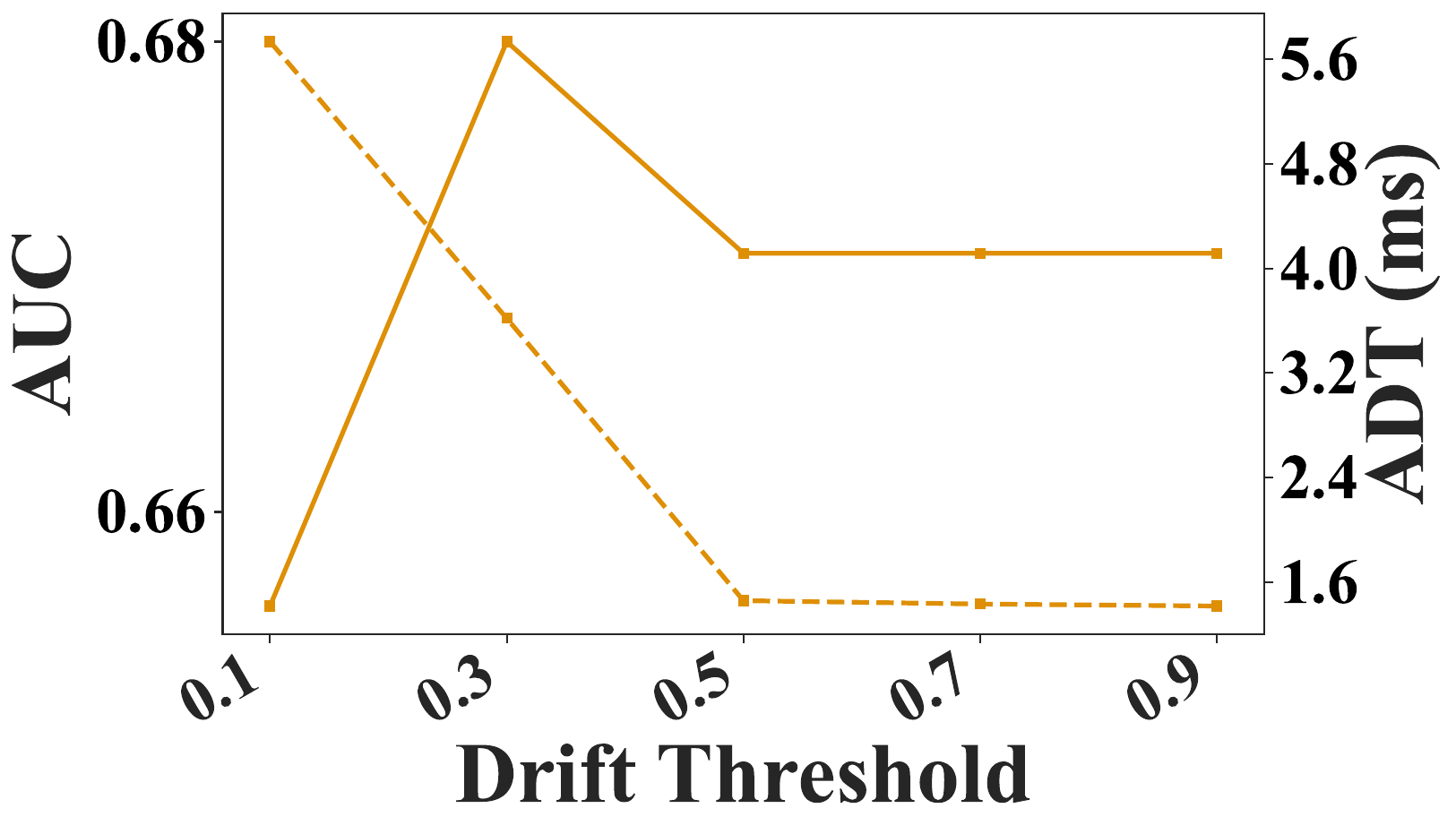}
        \caption{Dodgers.}
        \label{fig:dodgers_drift}
    \end{subfigure}
    \begin{subfigure}[b]{0.23\textwidth}
        \centering
        \includegraphics[width=\textwidth]{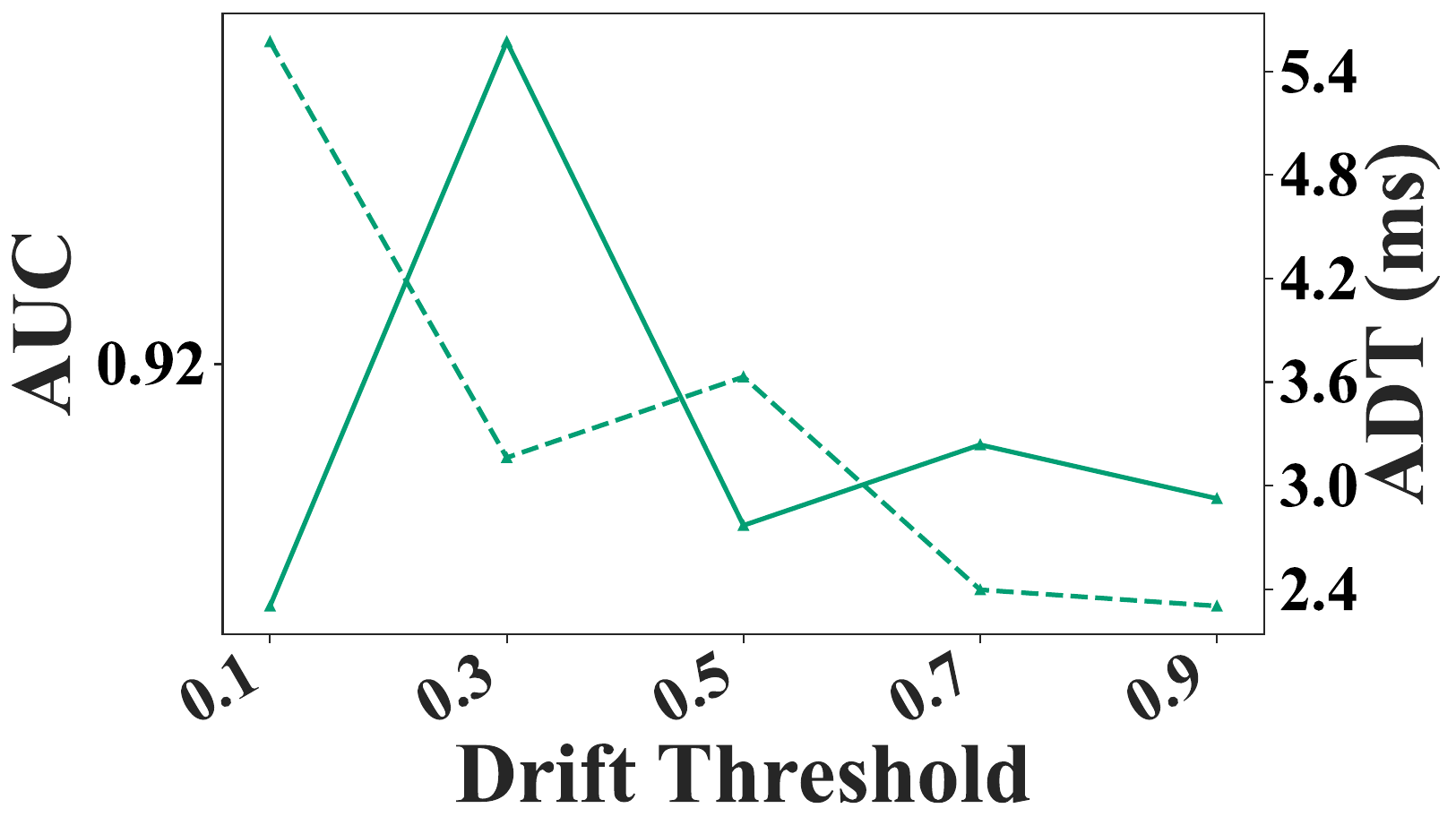}
        \caption{IOPS\_1c6.}
        \label{fig:motivation_iops}
    \end{subfigure}
    \begin{subfigure}[b]{0.23\textwidth}
        \centering
        \includegraphics[width=\textwidth]{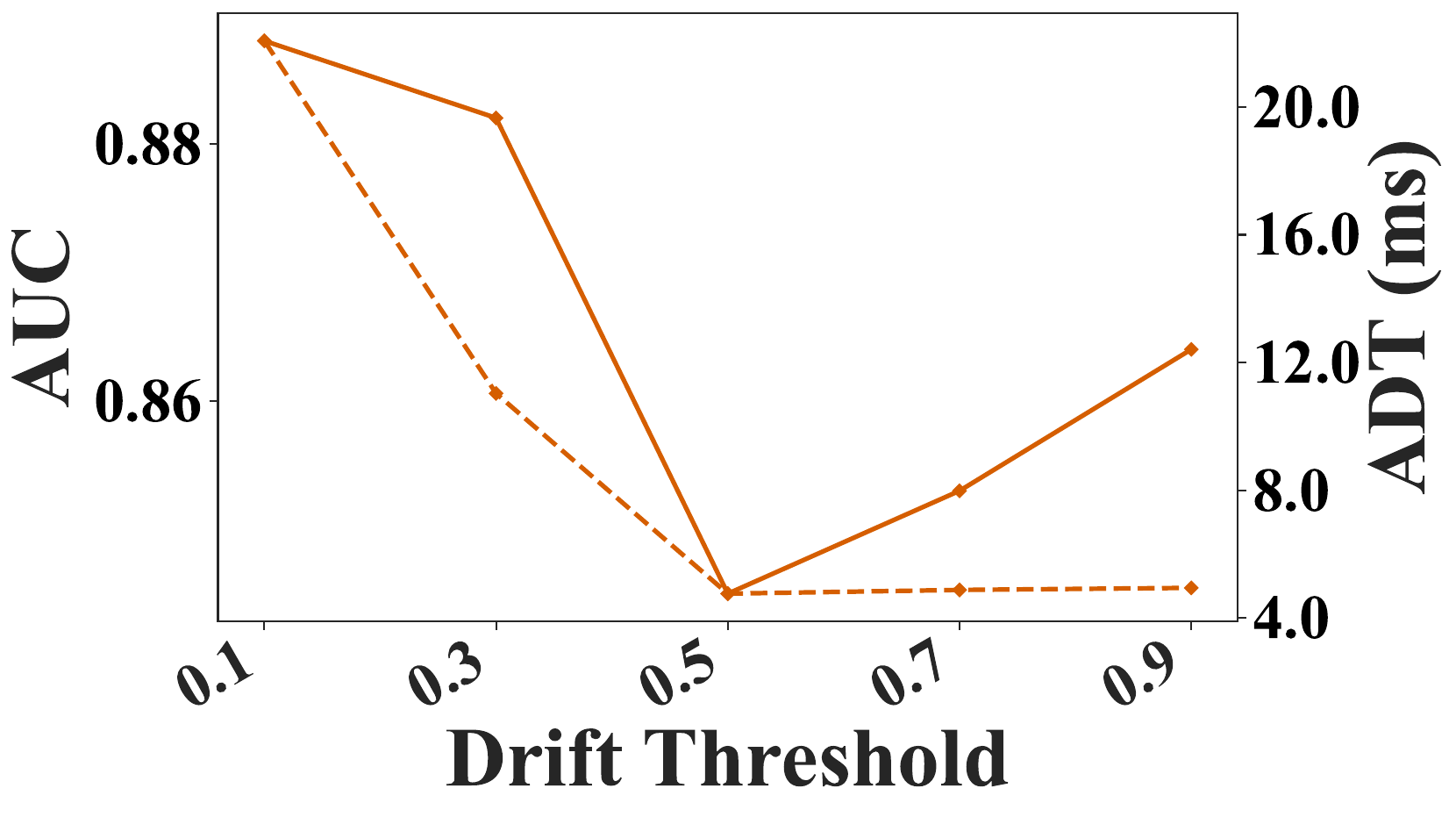}
        \caption{SMD\_1\_1.}
        \label{fig:motivation_smd}
    \end{subfigure}

    \caption{Evaluating the Influence of the Concept Drift Threshold on AUC and ADT.}
    \label{fig:drift_influence}
\end{figure*}

\begin{figure*}[htbp]
    \centering

    \begin{subfigure}[b]{0.8\textwidth}
        \centering
        \includegraphics[width=\textwidth]{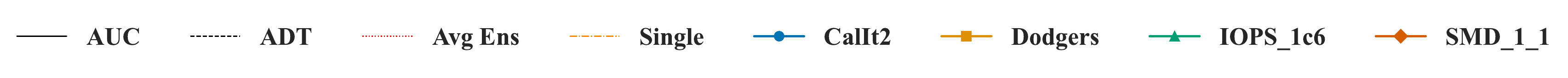}
    \end{subfigure}
    
    \begin{subfigure}[b]{0.23\textwidth}
        \centering
        \includegraphics[width=\textwidth]{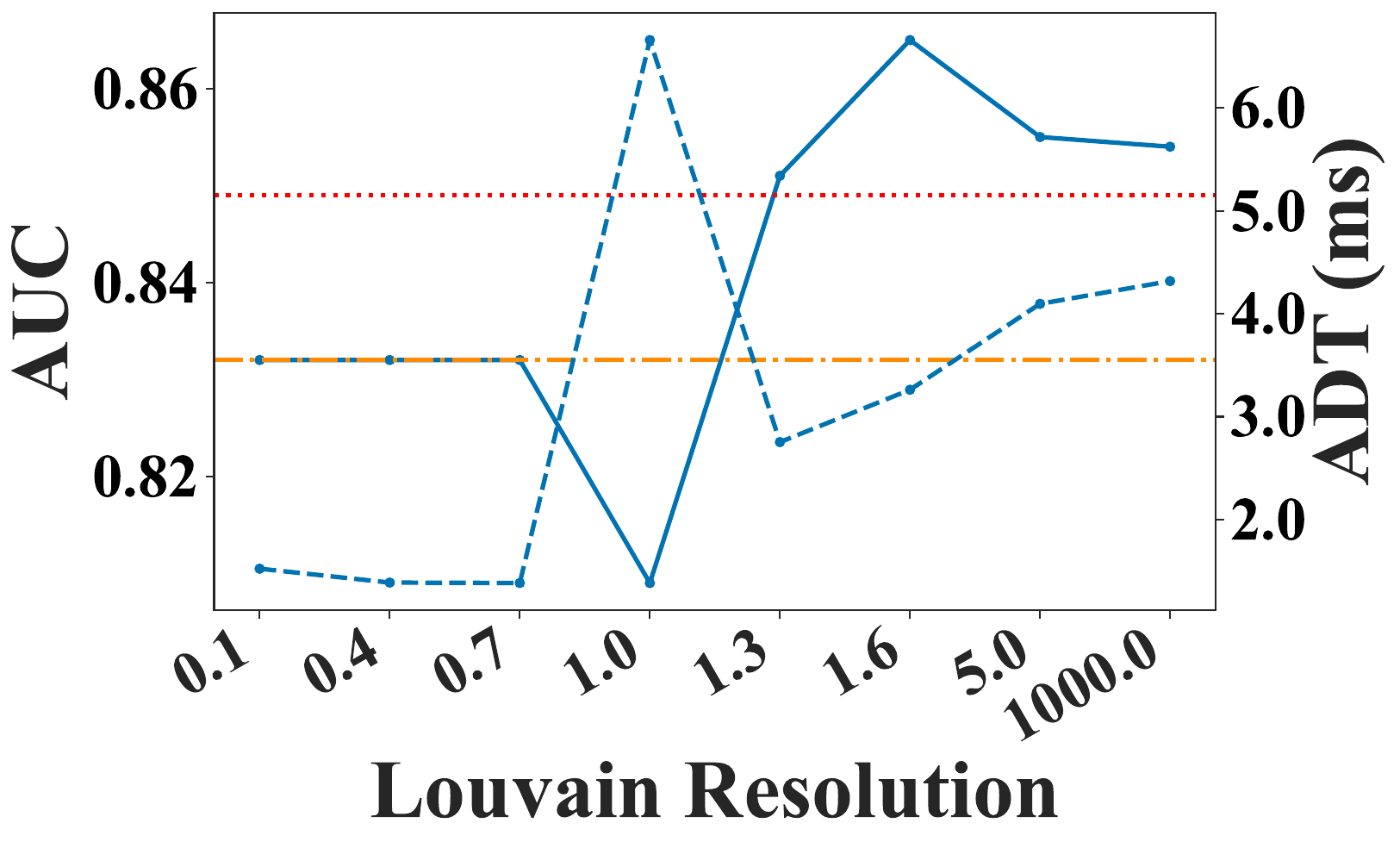}
        \caption{CalIt2.}
        \label{fig:calit2_drift}
    \end{subfigure}
    \begin{subfigure}[b]{0.23\textwidth}
        \centering
        \includegraphics[width=\textwidth]{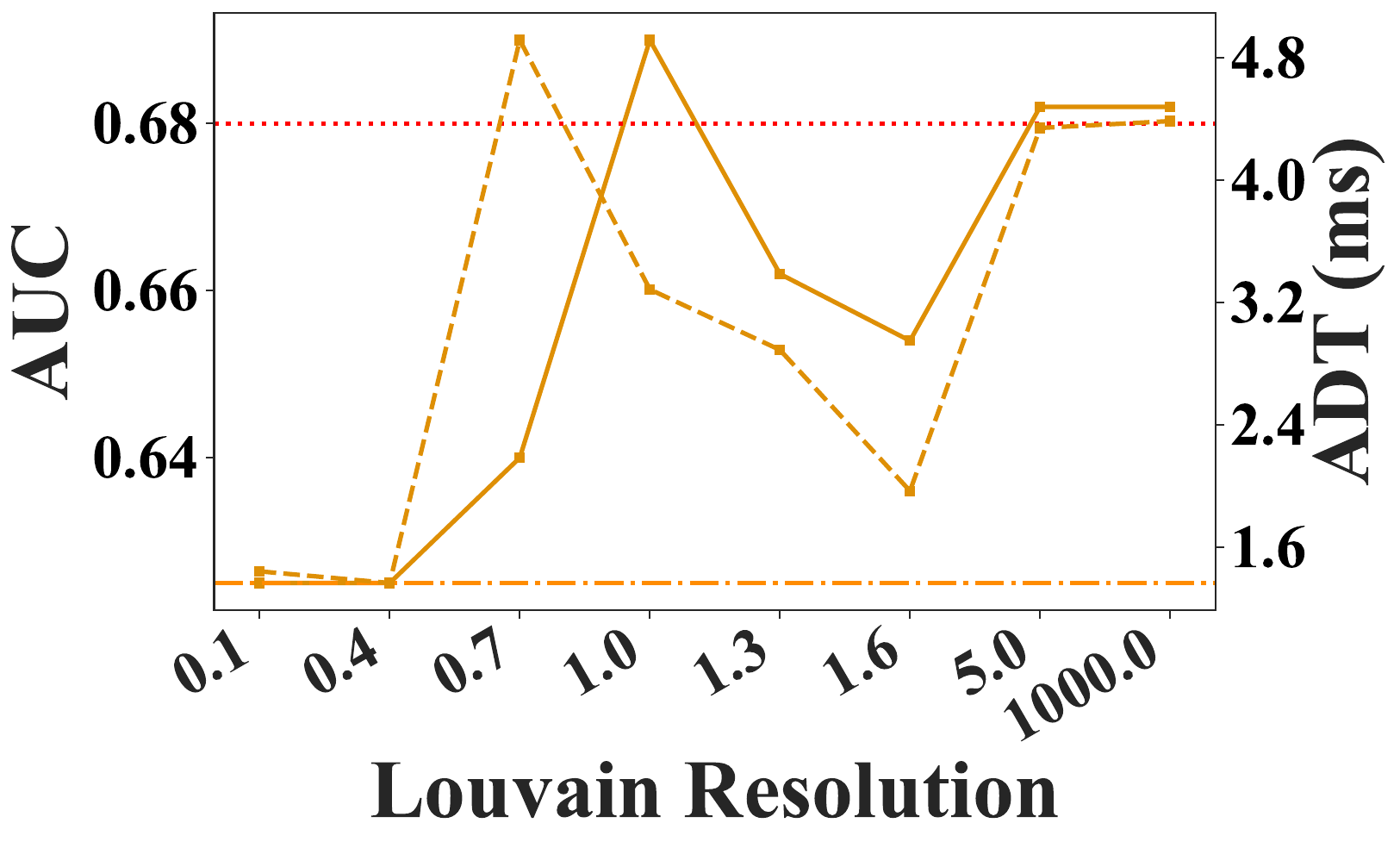}
        \caption{Dodgers.}
        \label{fig:dodgers_drift}
    \end{subfigure}
    \begin{subfigure}[b]{0.23\textwidth}
        \centering
        \includegraphics[width=\textwidth]{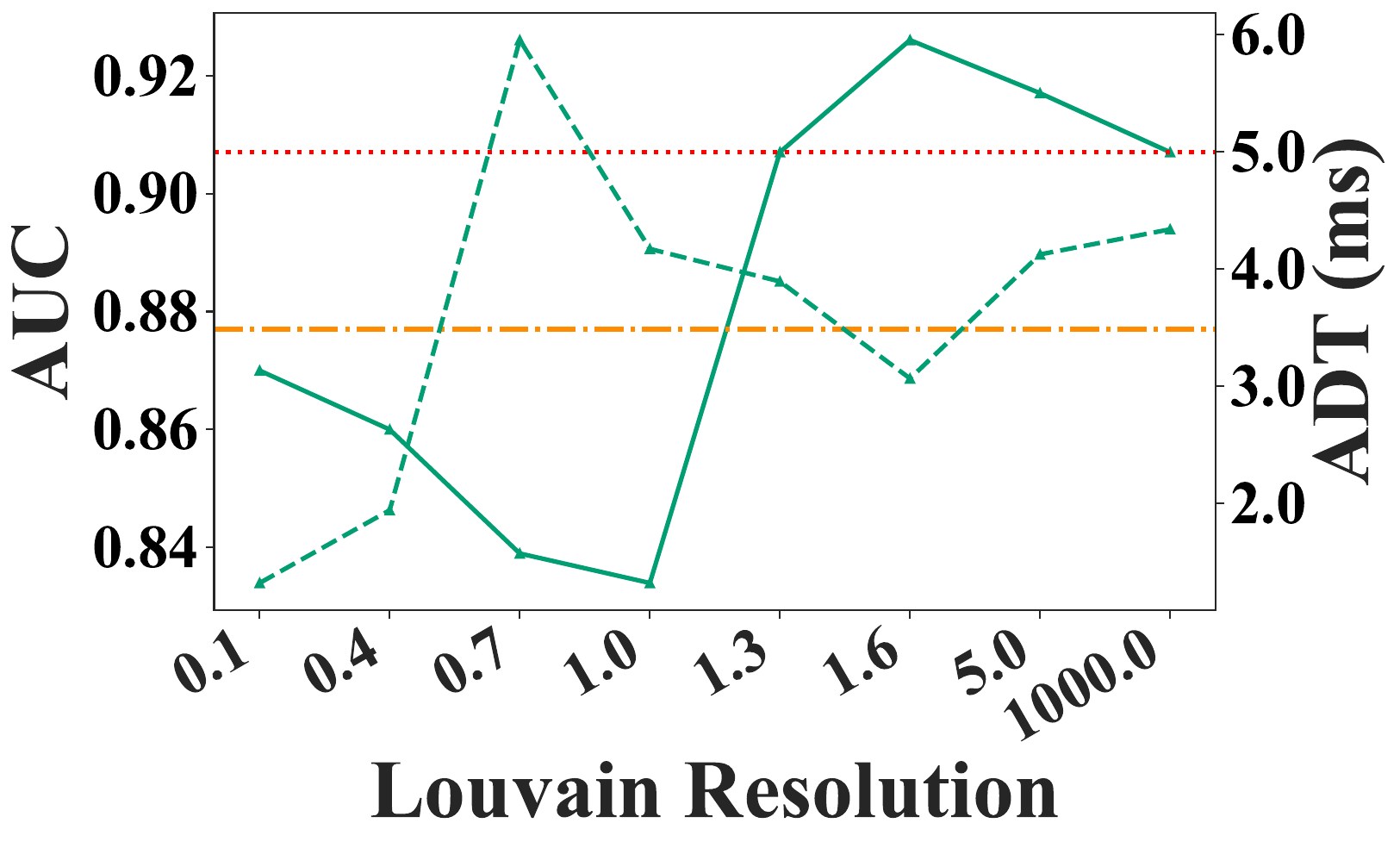}
        \caption{IOPS\_1c6.}
        \label{fig:motivation_iops}
    \end{subfigure}
    \begin{subfigure}[b]{0.23\textwidth}
        \centering
        \includegraphics[width=\textwidth]{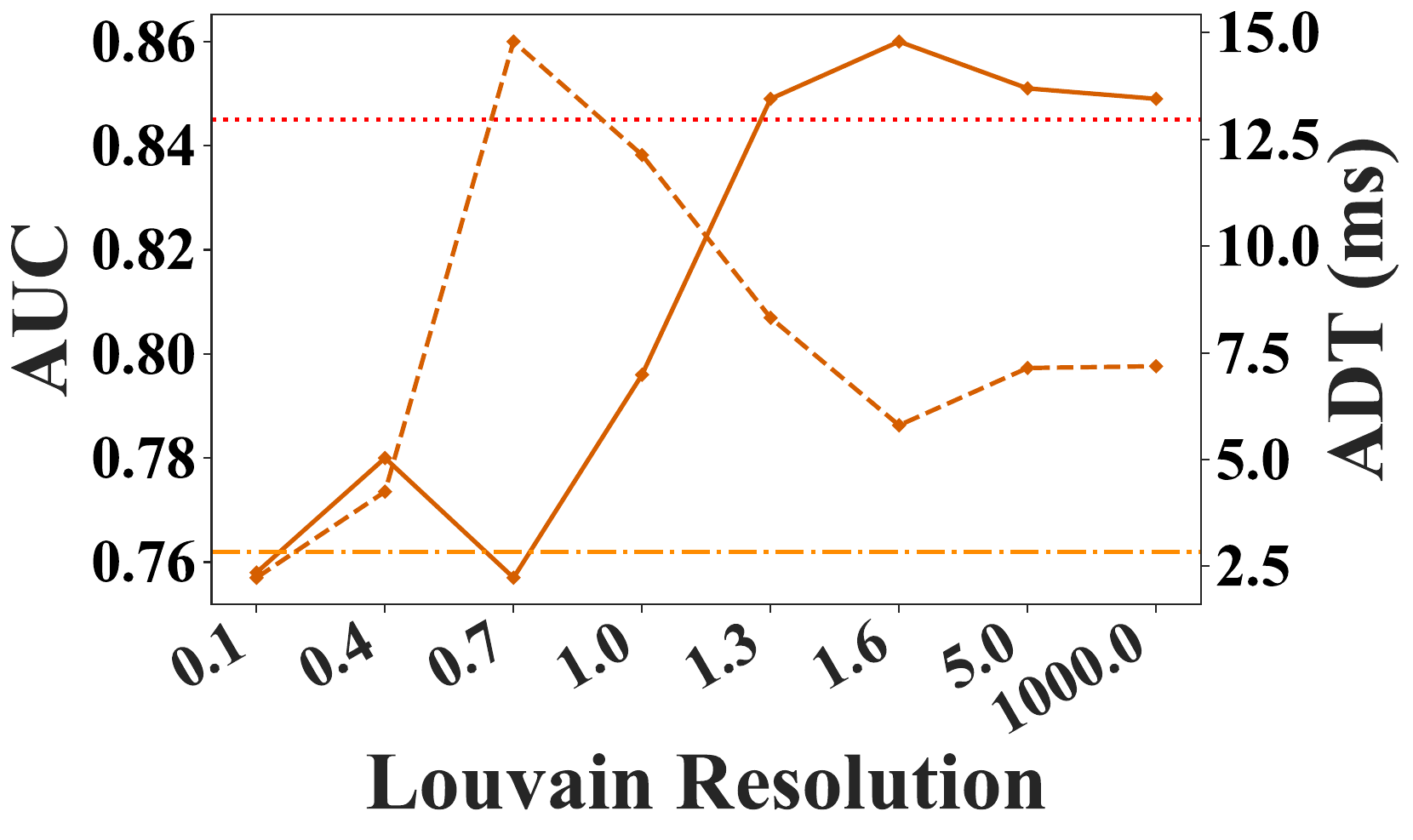}
        \caption{SMD\_1\_1.}
        \label{fig:motivation_smd}
    \end{subfigure}

    \caption{Evaluating the Influence of the Louvain Resolution Parameter on AUC and ADT.}
    \label{fig:lr_influence}
\end{figure*}

\subsection{Baseline Comparison}

Since some baselines run only on CPU, Table~\ref{tab:baseline_cmp} reports GDME with a CPU suffix for CPU implementation and without a suffix for GPU implementation. As shown in Table~\ref{tab:baseline_cmp}, GDME achieves the highest AUC across most datasets, e.g., 0.866 on CalIt2 ($\approx$24\% higher than MemStream, 0.695) and a 15\% improvement on SMD\_1\_1 compared with RCF. In terms of efficiency, GDME is slower than MemStream but more stable across datasets and substantially faster than RCF and xStream. Overall, GDME achieves a favorable balance between detection accuracy and efficiency, delivering significant AUC improvements while maintaining competitive detection times.

\subsection{Evaluation of Ensemble Effectiveness}

To evaluate the effectiveness of our ensemble strategy, we compared GDME with all algorithms it uses and with an average ensemble, as shown in Table~\ref{tab:ensemble_comparison}. Results show that GDME consistently outperforms any individual model in AUC, though with higher detection time. Compared with the average ensemble, GDME achieves higher AUC while using only 30\%--35\% of the detection time. These findings demonstrate that GDME effectively combines multiple models to capture diverse patterns while mitigating the influence of weaker models.

\subsection{Memory Usage Analysis}
\begin{figure}[htbp]
    \centering
    \includegraphics[width=0.9\linewidth]{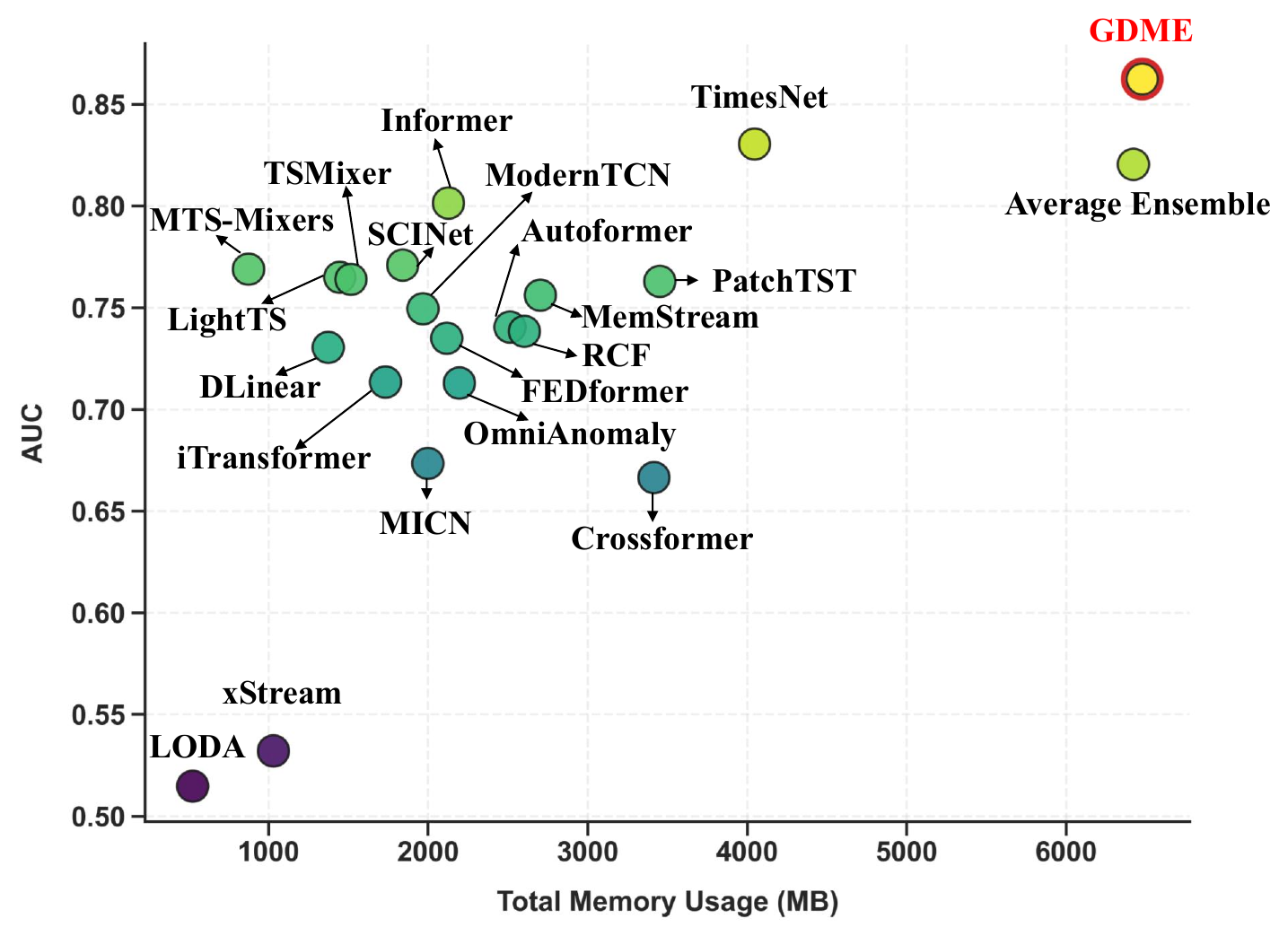}
    \caption{Memory–AUC Trade-off Across Methods.}
    \label{fig:memory_rank}
\end{figure}

Memory usage is an important aspect of computational efficiency. Figure.~\ref{fig:memory_rank} compares the memory consumption of different methods on the SMD dataset. Although our method uses more memory, it achieves higher AUC, highlighting the trade-off between resources and performance.

\subsection{Analysis of Core Framework Parameters}

\subsubsection{Drift Threshold}

The concept drift threshold ($\theta_{\text{drift}}$) controls the framework's sensitivity to distribution changes. A low threshold (e.g., $0.1$) leads to frequent drift detections, causing repeated model initializations and incremental training, which degrade both AUC and ADT. A higher value reduces drift detections, resulting in smaller variations in AUC and ADT and more stable performance. Empirically, thresholds between $0.2$ and $0.4$ balance accuracy and efficiency, capturing meaningful drifts while avoiding unnecessary updates, as shown in Fig.~\ref{fig:drift_influence}

\subsubsection{Louvain Resolution}
\label{sec:louvain_influence}

The Louvain resolution is a key hyperparameter in Louvain method, controlling the granularity of detected communities: higher values yield more, smaller communities, while lower values produce fewer, larger ones. Fig.~\ref{fig:lr_influence} shows its impact on AUC and ADT. A resolution between $1.0$ and $2.0$ balances effectiveness and efficiency. At very low resolution values, the framework approximates model selection on a single large community encompassing the entire graph, and the AUC curve closely follows the Single Model baseline. Conversely, at very high resolution values, most communities contain only a single model. Model selection is performed within each small community and outputs are aggregated, approximating a full model average ensemble. Accordingly, the AUC curve at high resolutions approaches the Average Ensemble baseline in figure.

\subsubsection{$\alpha$ and $\beta$}

\begin{figure}[htbp]
    \centering
    \includegraphics[width=0.64\linewidth]{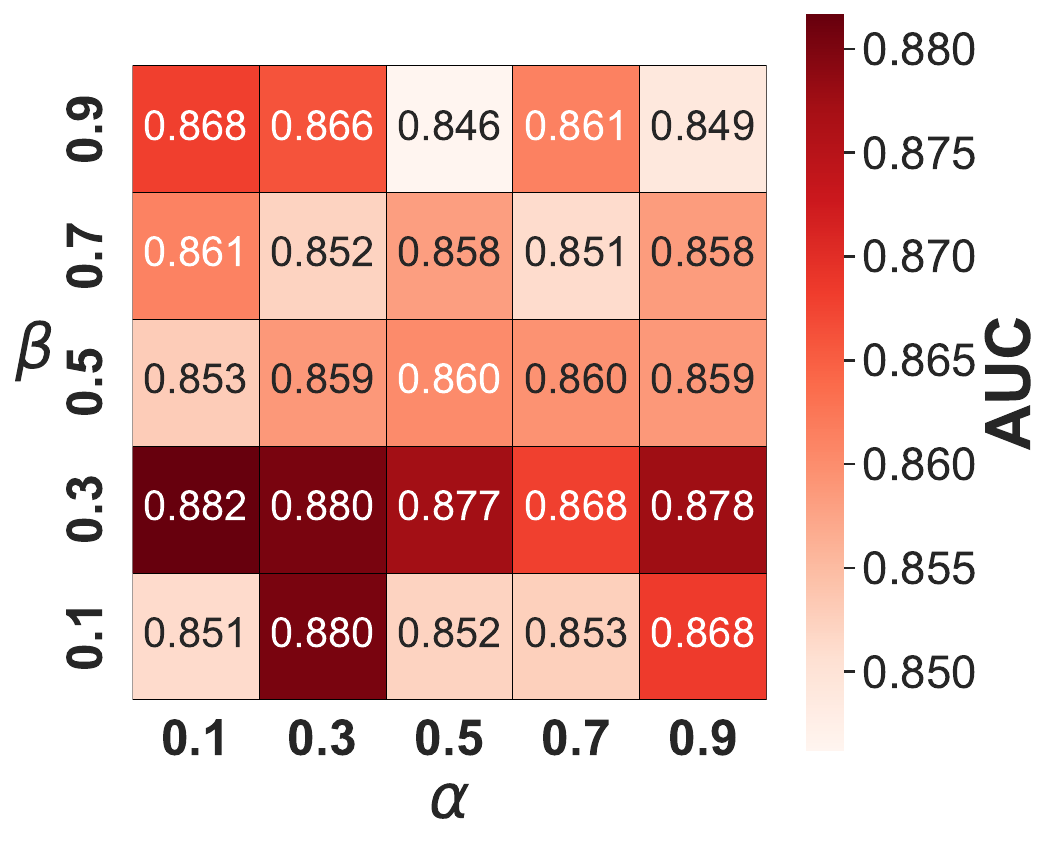}
    \caption{Impact of $\alpha$ and $\beta$ on AUC.}
    \label{fig:auc_heatmap_alpha_beta}
\end{figure}

Sections~\ref{sec:model_selection_for_ensemble} and~\ref{sec:graph_based_concept_drift_detection} introduce two parameters: $\alpha$, which balances centrality and pseudo-performance for representative model selection, and $\beta$, which balances community and centrality drift for concept drift detection. Experiments across multiple datasets suggest that the model generally performs well when $\alpha$ and $\beta$ lie between 0.3 and 0.7. Figure.~\ref{fig:auc_heatmap_alpha_beta} shows an example AUC heatmap on SMD\_1\_1 varying $\alpha$ and $\beta$.

\subsection{Ablation Study}

\subsubsection{Effectiveness of Community Partitioning}

\begin{figure}[htbp]
    \centering
    \hspace{8mm}
    \includegraphics[width=0.7\linewidth]{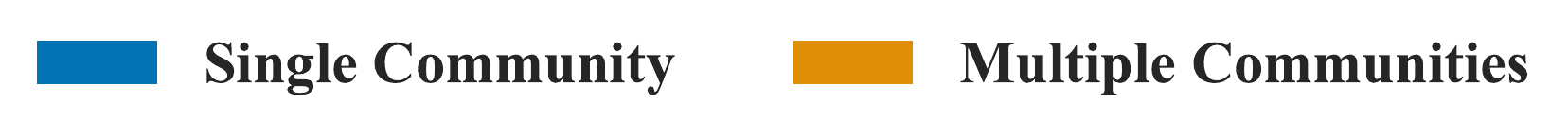}

    \includegraphics[width=0.8\linewidth]{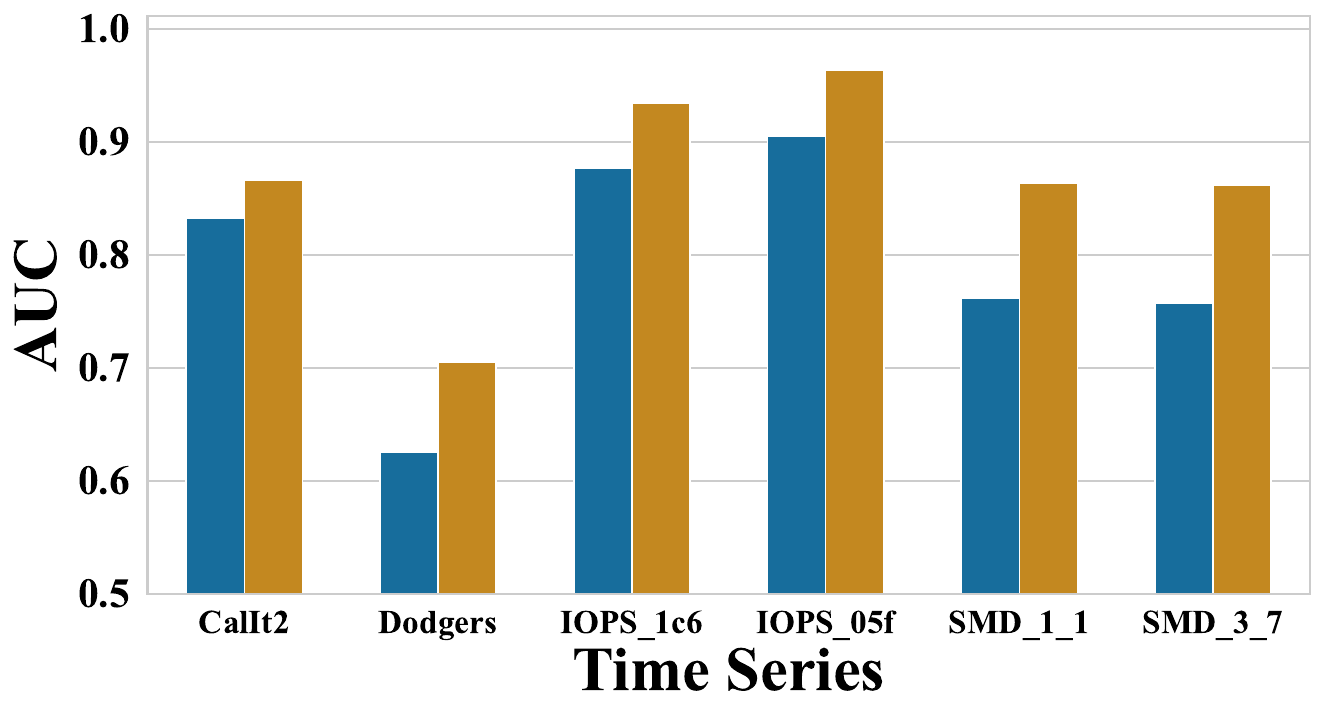}
    \caption{Comparison of community-based model selection and selection on the entire graph.}
    \label{fig:ablation_community_selection}
\end{figure}

As shown in Fig.~\ref{fig:ablation_community_selection}, we compare community-based model selection (Multiple Communities) with a strategy that treats the entire graph as a single community (Single Community). The results demonstrate that community-based selection consistently improves AUC across all datasets, with relative gains ranging from approximately 4\% to 14\%. This confirms that community-based model selection improves the robustness and diversity of the ensemble.

\subsubsection{Effectiveness of Selection Strategies}

\begin{figure}[htbp]
    \centering
    \hspace{8mm}
    \includegraphics[width=0.7\linewidth]{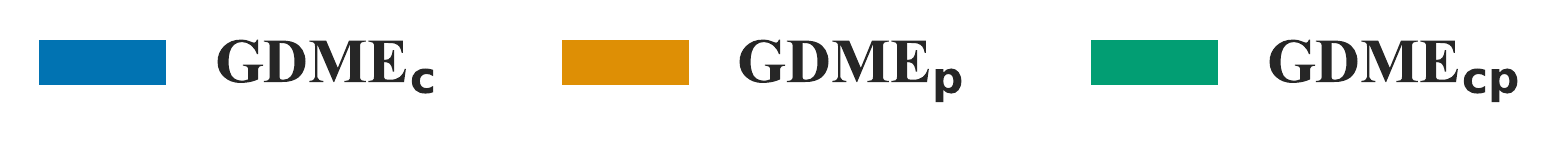}
    
    \includegraphics[width=0.8\linewidth]{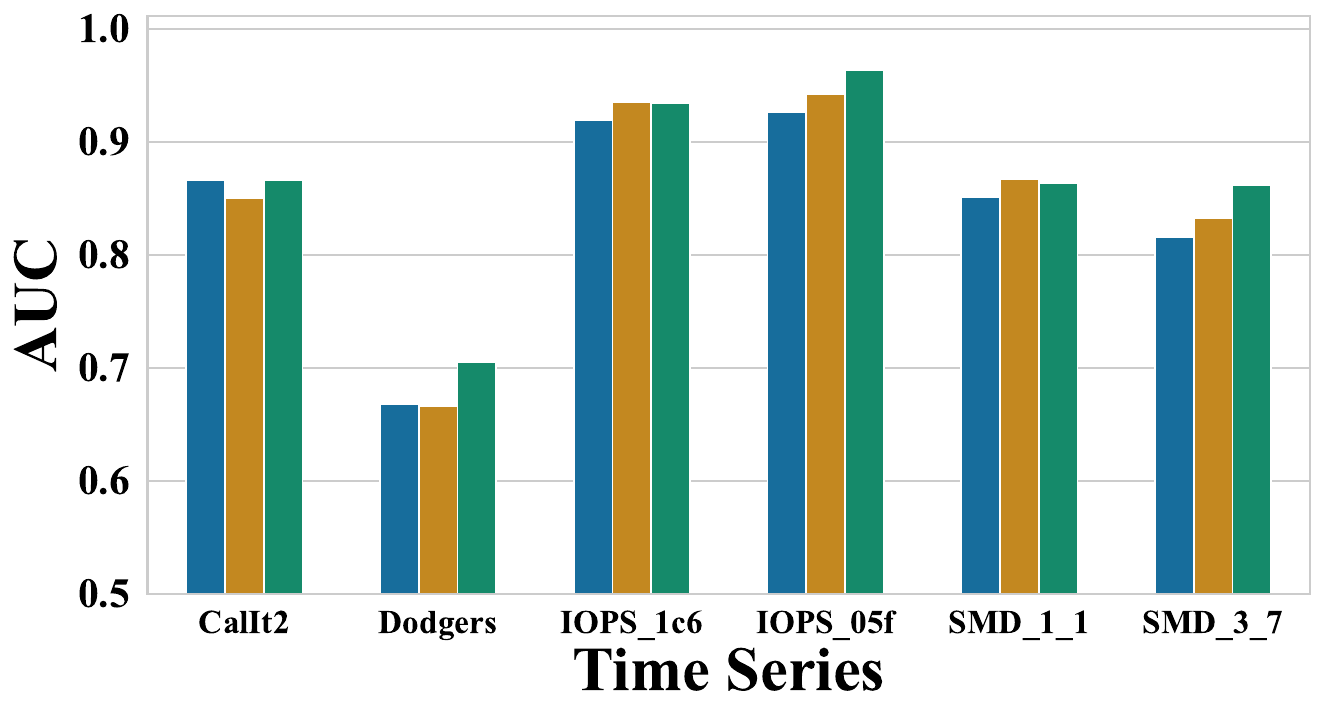}
    \caption{Comparison of model selection strategies within communities: $\text{GDME}_\text{c}$, $\text{GDME}_\text{p}$, and $\text{GDME}_\text{cp}$.}    
    \label{fig:ablation_selection_strategy}
\end{figure}

In this experiment, we evaluate three strategies for selecting models within each community: (1) centrality-based selection ($\text{GDME}_\text{c}$), (2) pseudo-performance-based selection ($\text{GDME}_\text{p}$), and (3) an equally weighted combination of the two ($\text{GDME}_\text{cp}$). As shown in Fig.~\ref{fig:ablation_selection_strategy}, comparing $\text{GDME}_\text{c}$ and $\text{GDME}_\text{p}$ with $\text{GDME}_\text{cp}$ shows that both individual strategies yield improvements on most datasets.

\section{CONCLUSION}
\label{sec:conclusion}

We propose GDME, a graph-based framework for online time series anomaly detection using model ensemble, achieving up to 24\% AUC improvement on seven heterogeneous datasets. Future work will explore additional graph-based structural information to further enhance ensemble performance.

\bibliographystyle{IEEEtran}
\bibliography{references}

@inproceedings{hundman2018detecting,
  title={Detecting spacecraft anomalies using lstms and nonparametric dynamic thresholding},
  author={Hundman, Kyle and Constantinou, Valentino and Laporte, Christopher and Colwell, Ian and Soderstrom, Tom},
  booktitle={Proceedings of the 24th ACM SIGKDD international conference on knowledge discovery \& data mining},
  pages={387--395},
  year={2018}
}

@inproceedings{su2019robust,
  title={Robust anomaly detection for multivariate time series through stochastic recurrent neural network},
  author={Su, Ya and Zhao, Youjian and Niu, Chenhao and Liu, Rong and Sun, Wei and Pei, Dan},
  booktitle={Proceedings of the 25th ACM SIGKDD international conference on knowledge discovery \& data mining},
  pages={2828--2837},
  year={2019}
}

@inproceedings{bhatia2022memstream,
  title={Memstream: Memory-based streaming anomaly detection},
  author={Bhatia, Siddharth and Jain, Arjit and Srivastava, Shivin and Kawaguchi, Kenji and Hooi, Bryan},
  booktitle={Proceedings of the ACM Web Conference 2022},
  pages={610--621},
  year={2022}
}

@inproceedings{guha2016robust,
  title={Robust random cut forest based anomaly detection on streams},
  author={Guha, Sudipto and Mishra, Nina and Roy, Gourav and Schrijvers, Okke},
  booktitle={International conference on machine learning},
  pages={2712--2721},
  year={2016},
  organization={PMLR}
}

@inproceedings{manzoor2018xstream,
  title={xstream: Outlier detection in feature-evolving data streams},
  author={Manzoor, Emaad and Lamba, Hemank and Akoglu, Leman},
  booktitle={Proceedings of the 24th ACM SIGKDD International Conference on Knowledge Discovery \& Data Mining},
  pages={1963--1972},
  year={2018}
}

@article{pevny2016loda,
  title={Loda: Lightweight on-line detector of anomalies},
  author={Pevn{\`y}, Tom{\'a}{\v{s}}},
  journal={Machine Learning},
  volume={102},
  number={2},
  pages={275--304},
  year={2016},
  publisher={Springer}
}

@article{paparrizos2022tsb,
  title={TSB-UAD: an end-to-end benchmark suite for univariate time-series anomaly detection},
  author={Paparrizos, John and Kang, Yuhao and Boniol, Paul and Tsay, Ruey S and Palpanas, Themis and Franklin, Michael J},
  journal={Proceedings of the VLDB Endowment},
  volume={15},
  number={8},
  pages={1697--1711},
  year={2022},
  publisher={VLDB Endowment}
}

@article{wang2024tssurvey,
  title={Deep Time Series Models: A Comprehensive Survey and Benchmark},
  author={Yuxuan Wang and Haixu Wu and Jiaxiang Dong and Yong Liu and Mingsheng Long and Jianmin Wang},
  booktitle={arXiv preprint arXiv:2407.13278},
  year={2024},
}

@article{aggarwal2015theoretical,
  title={Theoretical foundations and algorithms for outlier ensembles},
  author={Aggarwal, Charu C and Sathe, Saket},
  journal={Acm sigkdd explorations newsletter},
  volume={17},
  number={1},
  pages={24--47},
  year={2015},
  publisher={ACM New York, NY, USA}
}

@article{rayana2016less,
  title={Less is more: Building selective anomaly ensembles},
  author={Rayana, Shebuti and Akoglu, Leman},
  journal={Acm transactions on knowledge discovery from data (tkdd)},
  volume={10},
  number={4},
  pages={1--33},
  year={2016},
  publisher={ACM New York, NY, USA}
}

@inproceedings{caruana2006getting,
  title={Getting the most out of ensemble selection},
  author={Caruana, Rich and Munson, Art and Niculescu-Mizil, Alexandru},
  booktitle={Sixth International Conference on Data Mining (ICDM'06)},
  pages={828--833},
  year={2006},
  organization={IEEE}
}

@inproceedings{malhotra2015long,
  title={Long short term memory networks for anomaly detection in time series},
  author={Malhotra, Pankaj and Vig, Lovekesh and Shroff, Gautam and Agarwal, Puneet and others},
  booktitle={Proceedings},
  volume={89},
  number={9},
  pages={94},
  year={2015}
}

@inproceedings{zhou2021informer,
  title={Informer: Beyond efficient transformer for long sequence time-series forecasting},
  author={Zhou, Haoyi and Zhang, Shanghang and Peng, Jieqi and Zhang, Shuai and Li, Jianxin and Xiong, Hui and Zhang, Wancai},
  booktitle={Proceedings of the AAAI conference on artificial intelligence},
  volume={35},
  number={12},
  pages={11106--11115},
  year={2021}
}

@article{wu2021autoformer,
  title={Autoformer: Decomposition transformers with auto-correlation for long-term series forecasting},
  author={Wu, Haixu and Xu, Jiehui and Wang, Jianmin and Long, Mingsheng},
  journal={Advances in neural information processing systems},
  volume={34},
  pages={22419--22430},
  year={2021}
}

@inproceedings{zhou2022fedformer,
  title={Fedformer: Frequency enhanced decomposed transformer for long-term series forecasting},
  author={Zhou, Tian and Ma, Ziqing and Wen, Qingsong and Wang, Xue and Sun, Liang and Jin, Rong},
  booktitle={International conference on machine learning},
  pages={27268--27286},
  year={2022},
  organization={PMLR}
}

@inproceedings{
zhang2023crossformer,
title={Crossformer: Transformer Utilizing Cross-Dimension Dependency for Multivariate Time Series Forecasting},
author={Yunhao Zhang and Junchi Yan},
booktitle={International Conference on Learning Representations},
year={2023},
}

@inproceedings{Yuqietal-2023-PatchTST,
  title     = {A Time Series is Worth 64 Words: Long-term Forecasting with Transformers},
  author    = {Nie, Yuqi and
               H. Nguyen, Nam and
               Sinthong, Phanwadee and 
               Kalagnanam, Jayant},
  booktitle = {International Conference on Learning Representations},
  year      = {2023}
}

@inproceedings{
liu2024itransformer,
title={iTransformer: Inverted Transformers Are Effective for Time Series Forecasting},
author={Yong Liu and Tengge Hu and Haoran Zhang and Haixu Wu and Shiyu Wang and Lintao Ma and Mingsheng Long},
booktitle={The Twelfth International Conference on Learning Representations},
year={2024},
}

@inproceedings{zeng2023transformers,
  title={Are transformers effective for time series forecasting?},
  author={Zeng, Ailing and Chen, Muxi and Zhang, Lei and Xu, Qiang},
  booktitle={Proceedings of the AAAI conference on artificial intelligence},
  volume={37},
  number={9},
  pages={11121--11128},
  year={2023}
}

@article{Zhang2022LessIM,
  title={Less Is More: Fast Multivariate Time Series Forecasting with Light Sampling-oriented MLP Structures},
  author={T. Zhang and Yizhuo Zhang and Wei Cao and J. Bian and Xiaohan Yi and Shun Zheng and Jian Li},
  journal={ArXiv},
  year={2022},
  volume={abs/2207.01186}
}

@article{
chen2023tsmixer,
title={{TSM}ixer: An All-{MLP} Architecture for Time Series Forecast-ing},
author={Si-An Chen and Chun-Liang Li and Sercan O Arik and Nathanael Christian Yoder and Tomas Pfister},
journal={Transactions on Machine Learning Research},
issn={2835-8856},
year={2023}
}

@article{Li2023MTSMixersMT,
  title={MTS-Mixers: Multivariate Time Series Forecasting via Factorized Temporal and Channel Mixing},
  author={Zhe Li and Zhongwen Rao and Lujia Pan and Zenglin Xu},
  journal={ArXiv},
  year={2023},
  volume={abs/2302.04501}
}

@inproceedings{
wang2023micn,
title={{MICN}: Multi-scale Local and Global Context Modeling for Long-term Series Forecasting},
author={Huiqiang Wang and Jian Peng and Feihu Huang and Jince Wang and Junhui Chen and Yifei Xiao},
booktitle={The Eleventh International Conference on Learning Representations },
year={2023}
}

@article{liu2022scinet,
  title={Scinet: Time series modeling and forecasting with sample convolution and interaction},
  author={Liu, Minhao and Zeng, Ailing and Chen, Muxi and Xu, Zhijian and Lai, Qiuxia and Ma, Lingna and Xu, Qiang},
  journal={Advances in Neural Information Processing Systems},
  volume={35},
  pages={5816--5828},
  year={2022}
}

@inproceedings{
wu2023timesnet,
title={TimesNet: Temporal 2D-Variation Modeling for General Time Series Analysis},
author={Haixu Wu and Tengge Hu and Yong Liu and Hang Zhou and Jianmin Wang and Mingsheng Long},
booktitle={The Eleventh International Conference on Learning Representations },
year={2023}
}

@inproceedings{
donghao2024moderntcn,
title={Modern{TCN}: A Modern Pure Convolution Structure for General Time Series Analysis},
author={Luo donghao and wang xue},
booktitle={The Twelfth International Conference on Learning Representations},
year={2024},
}

@inproceedings{yoon2022adaptive,
  title={Adaptive model pooling for online deep anomaly detection from a complex evolving data stream},
  author={Yoon, Susik and Lee, Youngjun and Lee, Jae-Gil and Lee, Byung Suk},
  booktitle={Proceedings of the 28th ACM SIGKDD conference on knowledge discovery and data mining},
  pages={2347--2357},
  year={2022}
}

@article{gama2013evaluating,
  title={On evaluating stream learning algorithms},
  author={Gama, Joao and Sebastiao, Raquel and Rodrigues, Pedro Pereira},
  journal={Machine learning},
  volume={90},
  number={3},
  pages={317--346},
  year={2013},
  publisher={Springer}
}

@article{goswami2022unsupervised,
  title={Unsupervised model selection for time-series anomaly detection},
  author={Goswami, Mononito and Challu, Cristian and Callot, Laurent and Minorics, Lenon and Kan, Andrey},
  journal={arXiv preprint arXiv:2210.01078},
  year={2022}
}

@article{blondel2008fast,
  title={Fast unfolding of communities in large networks},
  author={Blondel, Vincent D and Guillaume, Jean-Loup and Lambiotte, Renaud and Lefebvre, Etienne},
  journal={Journal of statistical mechanics: theory and experiment},
  volume={2008},
  number={10},
  pages={P10008},
  year={2008},
  publisher={IOP Publishing}
}

@inproceedings{ihler2006adaptive,
  title={Adaptive event detection with time-varying poisson processes},
  author={Ihler, Alexander and Hutchins, Jon and Smyth, Padhraic},
  booktitle={Proceedings of the 12th ACM SIGKDD international conference on Knowledge discovery and data mining},
  pages={207--216},
  year={2006}
}

@inproceedings{ren2019time,
  title={Time-series anomaly detection service at microsoft},
  author={Ren, Hansheng and Xu, Bixiong and Wang, Yujing and Yi, Chao and Huang, Congrui and Kou, Xiaoyu and Xing, Tony and Yang, Mao and Tong, Jie and Zhang, Qi},
  booktitle={Proceedings of the 25th ACM SIGKDD international conference on knowledge discovery \& data mining},
  pages={3009--3017},
  year={2019}
}

@inproceedings{tong2016online,
  title={Online mobile micro-task allocation in spatial crowdsourcing},
  author={Tong, Yongxin and She, Jieying and Ding, Bolin and Wang, Libin and Chen, Lei},
  booktitle={2016 IEEE 32Nd international conference on data engineering (ICDE)},
  pages={49--60},
  year={2016},
  organization={IEEE}
}

\end{document}